\documentclass[camera]{jpaper}

\usepackage[normalem]{ulem}
\usepackage{amsmath}
\usepackage{comment}
\usepackage{xcolor}         
\usepackage{mathtools,nccmath}
\usepackage{algorithm}
\usepackage{algorithmic}
\usepackage{amsmath}

\DeclareRobustCommand{\bigO}{
  \text{\usefont{OMS}{cmsy}{m}{n}O}
}

\floatname{algorithm}{Function}

\newcommand{\ours}{\texttt{OLLA}}

\begin{document}

\title{{\LARGE \ours{}}: Optimizing the Lifetime and Location of Arrays\\to Reduce the Memory Usage of Neural Networks}

\author{
Benoit Steiner\thanks{Correspondance to benoitsteiner@meta.com} \\ (FAIR) \and 
Mostafa Elhoushi \\(Meta) \and
Jacob Kahn \\ (FAIR) \and
James Hegarty \\ (Meta)
}

\date{}
\maketitle

\thispagestyle{empty}

\begin{abstract}

The size of deep neural networks has grown exponentially in recent years. Unfortunately, hardware devices have not kept pace with the rapidly increasing memory requirements. To cope with this, researchers have turned to techniques such as spilling and recomputation, which increase training  time, or reduced precision and model pruning, which can affect model accuracy.

We present \ours{}, an algorithm that optimizes the lifetime and memory location of the tensors used to train neural networks. Our method reduces the memory usage of existing neural networks, without needing any modification to the models or their training procedures.


We formulate the problem as a joint integer linear program (ILP). We present several techniques to simplify the encoding of the problem, and enable our approach to scale to the size of  state-of-the-art neural networks using an off-the-shelf ILP solver. We experimentally demonstrate that \ours{} only takes minutes if not seconds to allow the training of neural networks using one-third less memory on average. \ours{} is available at https://github.com/facebookresearch/OLLA.

\end{abstract}

\section{Introduction}

\begin{figure}
    \centering
    \includegraphics[width=0.475\textwidth]{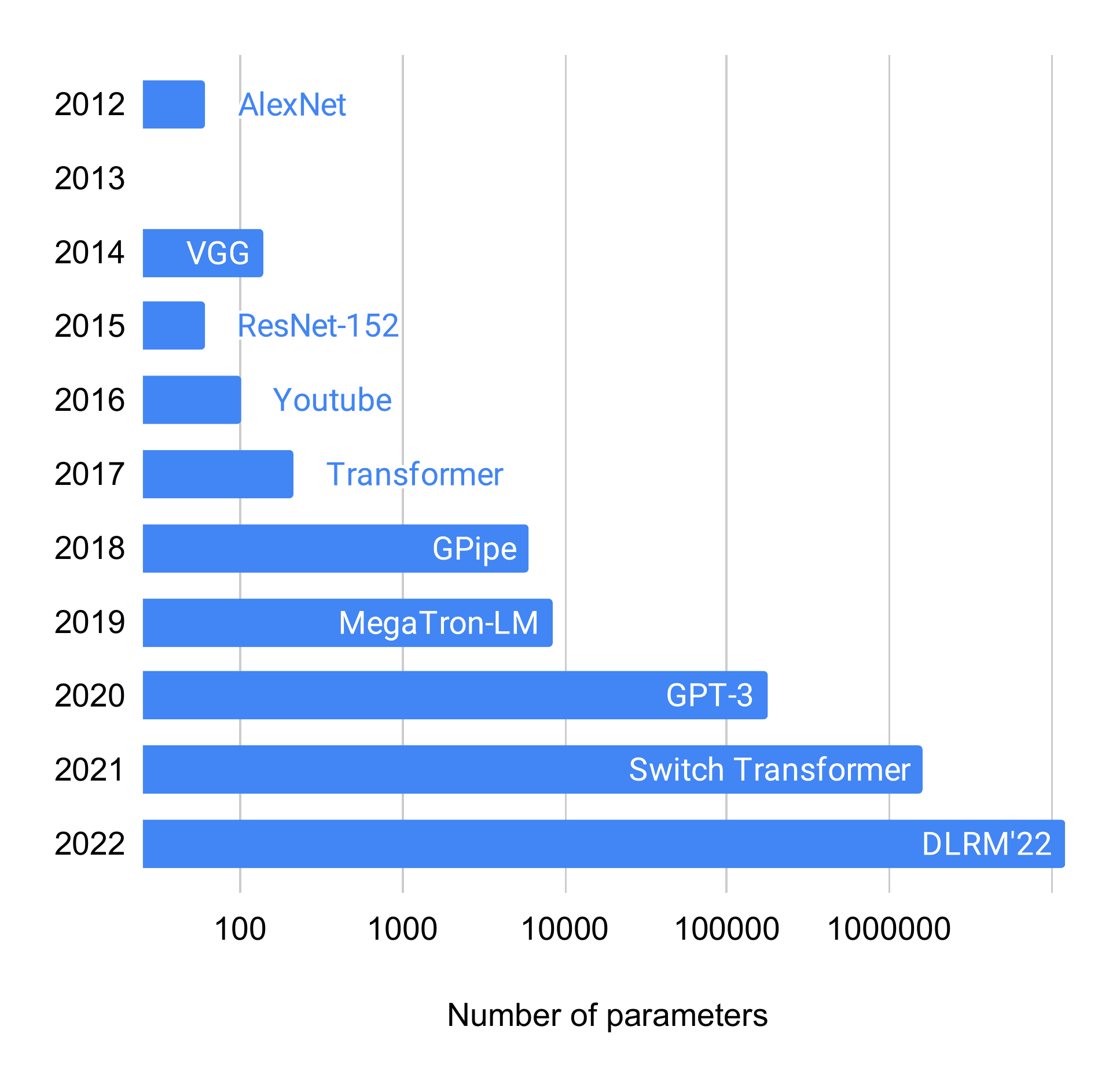}
    \caption{The number of deep neural network parameters has increased by 100,000 fold over the last 10 years, starting to grow exponentially around 2016. The x-axis is plotted on a log scale.}
    \label{fig:dnn_size_over_time}
\end{figure}

Scale is a major force behind the accuracy improvements of machine-learning-based solutions~\cite{RobustnessOverParameterization}, and both the depth and width of deep neural networks (DNN) are expanding exponentially~\cite{dnn_trends} (Figure~\ref{fig:dnn_size_over_time}). This inflation in size increases the memory needed to store the weights of the neural network and the intermediate results (e.g., activations and gradients) generated during the training process. Compounding the problem, researchers are training neural networks on larger inputs, such as high-resolution images~\cite{superresolution, Tai_2017_CVPR}, video~\cite{SlowFast}, three dimensional point-clouds~\cite{Chen2017Multiview3O}, long natural language sequences~\cite{transformer, sparse_transformers, bert}, and using larger batch sizes to increase efficiency~\cite{increaseTheBatchSize}.

\paragraph{}

\begin{figure}
    \centering
    \includegraphics[width=0.475\textwidth]{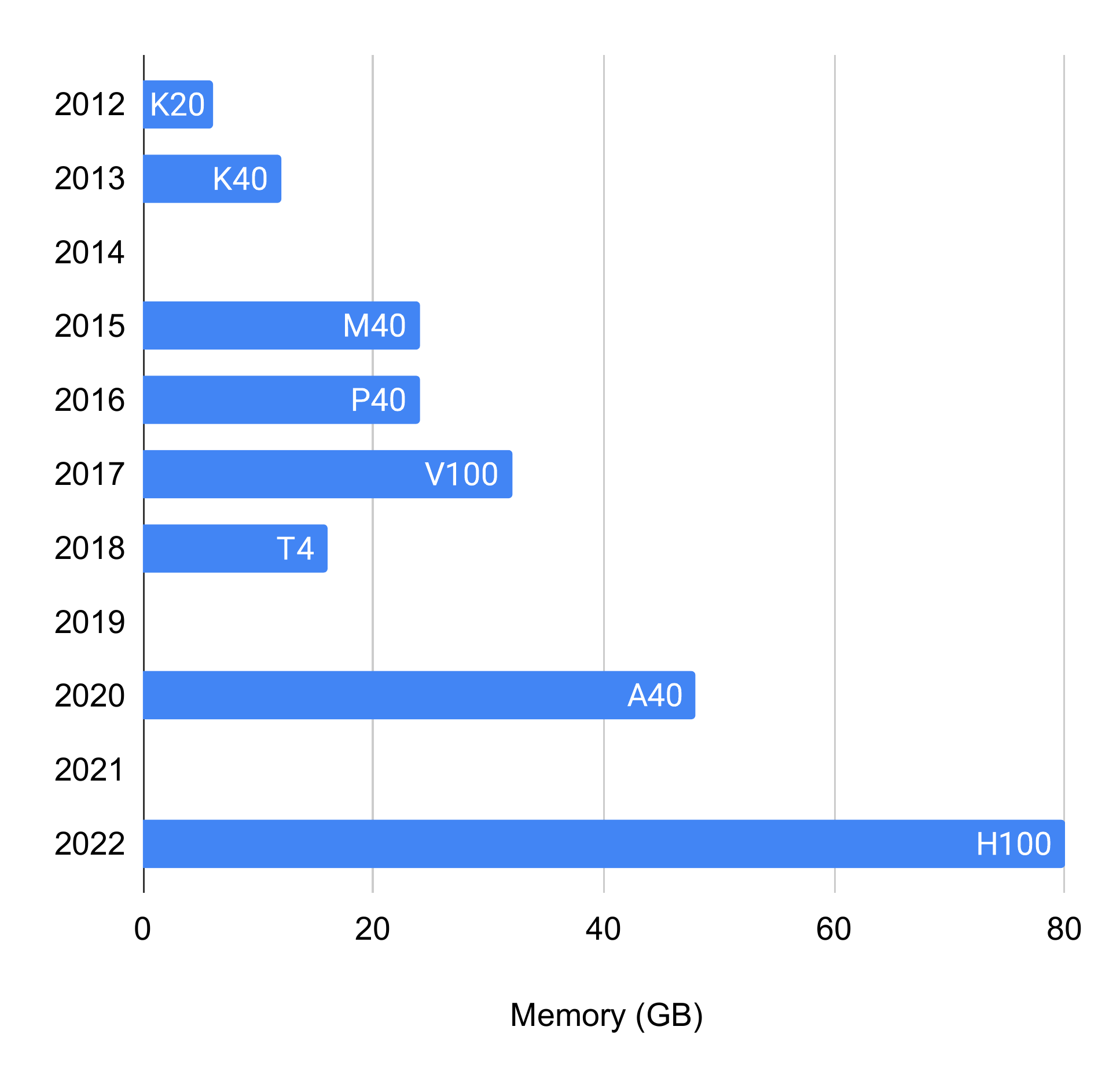}
    \caption{The memory capacity of NVidia datacenter GPUs (in gigabytes) has only increased tenfold over the last decade, which has not kept pace with the rapidly increasing size of deep neural networks. The x-axis is plotted on a linear scale.}
    \label{fig:gpu_memory_over_time}
\end{figure}

Unfortunately, due to the slowing of Moore’s law, the memory capacity of hardware has only increased linearly over the last decade (Figure~\ref{fig:gpu_memory_over_time}). Thus, the amount of memory available on the hardware used to train DNNs has not kept pace with the needs of deep learning. Furthermore, features powered by machine learning, such as automatic speech recognition~\cite{paulik} or keyboard suggestions~\cite{Hard2018FederatedLF}, are being personalized by fine tuning models on-device. This means that model training is increasingly being pushed to even more memory constrained edge devices such as smartphones. As a result, memory is increasingly becoming a bottleneck that hinders progress, and researchers frequently mention memory scarcity as a limiting factor that impacts their work~\cite{alexnet, resnet, Semantic_Segmentation, transformerXL, sparse_transformers}.

\paragraph{}

The research community has proposed several solutions to mitigate the problem. Data spilling~\cite{SwapOutSwapIn} and recomputation of intermediate results~\cite{checkmate} relieve memory pressure. Novel neural network architectures~\cite{mobilenet} use memory more sparingly. Strategies such as reduced precision training~\cite{fp8Training,INT8Training,bfloat16Training} and weight pruning~\cite{TaylorPruning,elkerdawy2022fire,He_2020_CVPR} decrease memory requirements. However, these all come at the cost of decreasing the accuracy of the model, or increasing the time it takes to train it, or both~\cite{LostInPruning,ChallengesTransformerQuantization,OscilationsQAT}.

\paragraph{}

Popular deep learning frameworks such as PyTorch~\cite{PyTorch} and TensorFlow~\cite{tensorflow} do not fully utilize the limited memory
available. Similar to traditional dynamic memory allocators such as tcmalloc~\cite{TCMalloc} and jemalloc~\cite{jemalloc}, these frameworks maintain a pool of free blocks of memory at runtime. To serve memory requests, they look for a large enough memory block in the memory pool, or allocate it from the physical memory if none is available. This results in memory fragmentation when free memory blocks do not exactly match the size of an allocation request, which occurs frequently.

\paragraph{}

Furthermore, DNN frameworks do not optimize tensor lifetimes. PyTorch~\cite{PyTorch} executes operations in the order in which they are defined in the program. TensorFlow~\cite{tensorflow} keeps a queue of operators that are ready to run, and executes them on a first-come, first-served basis. As a result, tensors can be allocated earlier than required, or freed later than necessary, wasting valuable memory.

\paragraph{}

\begin{figure}
    \centering
    \includegraphics[width=0.475\textwidth]{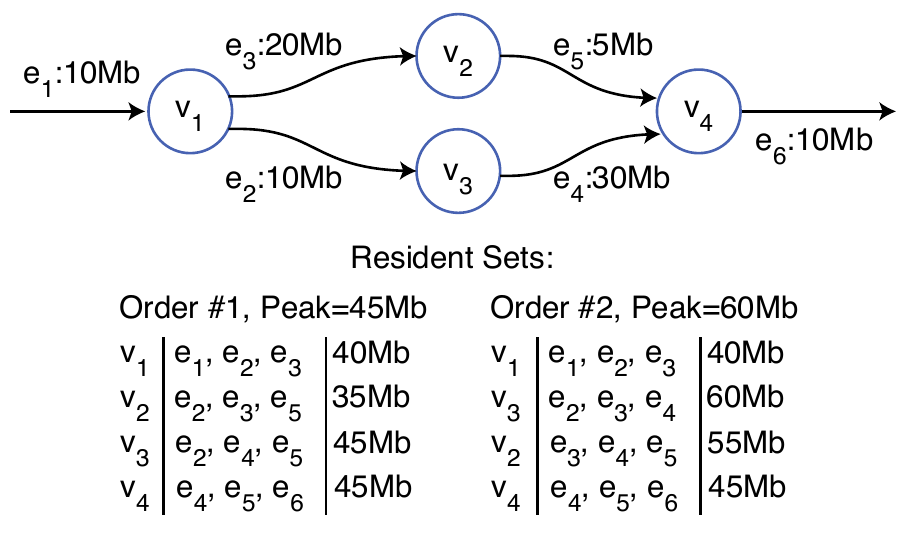}
    \caption{Node execution orders can impact peak memory usage. Edges are annotated with the size of their corresponding tensors, and the two feasible node orders are annotated with the set of edges resident in memory at each step. Running $v_2$ before $v_3$ is significantly more memory efficient.}
    \label{fig:scheduling}
\end{figure}

Our method overcomes these two limitations of existing deep learning frameworks. We model the computations performed to train a deep neural network as a dataflow graph of operations. We analyze this graph to find a topological ordering of the nodes that adjusts the lifetime of the tensors generated by these operations to minimize the peak amount of memory that needs to be allocated (Figure~\ref{fig:scheduling}). Furthermore, we find an optimal packing of these tensors, which minimizes memory fragmentation (Figure~\ref{fig:layoutopt}). We encode these two objectives as an integer linear program (ILP) that can be solved quickly by commodity solvers, and present \ours{} (Optimization of the Lifetime and Location of Arrays), our algorithm for memory-optimal training of neural networks.

\paragraph{}

In addition to significantly reducing memory usage, our solution has four key strengths. First, it does not impact the accuracy of the predictions of the neural networks. Second, it requires no modification to the neural network or the training procedure. Third, it doesn't increase training time. Fourth, it is orthogonal to and can be combined with other memory reductions techniques to further reduce the memory needs of a neural network.

\paragraph{}

Our work makes the following novel contributions:
\begin{itemize}
    \item We formulate the problem of finding the lifetime and memory location of tensors that minimizes the peak memory required to train neural networks as a joint integer linear program.
    \item We demonstrate how to leverage domain knowledge to simplify the ILP formulation, which enables off-the-shelf solvers to quickly reduce the memory usage of large DNNs.
    \item We study empirically the practicality and effectiveness of our solution on a wide variety of DNNs, which achieves average memory savings exceeding 30\% in a median time of less than 10 seconds.
    \item We provide an open source implementation of \ours{}, which is available at https://github.com/facebookresearch/OLLA.
\end{itemize}

\section{Background}

\subsection{Representing Neural Networks as Dataflow Graphs}
\label{dag}
Deep neural networks can be represented using dataflow graphs, as pioneered by TensorFlow~\cite{tensorflow}. The nodes of the graph encode the computations to be performed (e.g. matrix multiplications, convolutions, activation functions), while the edges represent the data (\textit{aka} tensor or array) that is produced by an operation and transferred to consumer nodes.

\paragraph{}

Due to the producer-consumer relation between connected nodes, edges are oriented. Each edge has exactly one source, which is the operator that generated the corresponding tensor. Since a tensor can be consumed by more than one node, edges can have multiple sinks.

\paragraph{}

Operators can have multiple incoming (\textit{aka} fanin) edges. Typically, one of these incoming edges will be the tensor generated by the previous layer, and another one will be a weight tensor. Similarly, operators can have multiple outgoing (\textit{aka} fanout) edges: while most operations generate a single output tensor, some may create two or more. Operators with no fanout edges are used to model the final outputs of the neural network. Operators without fanin edges can model random number generators, constants, weights, or initial inputs to the neural network. 

\paragraph{}

In the remainder of this paper, we assume that the graphs are acyclic. In practice, this is not a significant limitation since recurrent neural networks such as LSTM~\cite{hochreiter1997lstm} have been eclipsed by transformers~\cite{transformer}. Furthermore, their loops can be unrolled to avoid the problem altogether.

\subsection{Optimizing Tensor Lifetimes}

In order for an operator to run, all its input tensors must be resident in memory, and its output tensors must have been allocated so that they can be written to while the node executes. Additionally, to avoid recomputing tensors, once a tensor is generated it must be preserved in memory until all its consumers have been run. 

\paragraph{}

We define the resident set $RS(t)$ at a given time $t$ as the set of tensors that need to be kept in memory at that point in the execution of the neural network. It comprises the tensors in the fanin and fanout of the operator that is scheduled for execution at timestep $t$, as well as all the other tensors that were previously generated but need to be kept in memory to be able to run subsequent operators. The peak resident set is the largest resident set over the execution of the network.

\paragraph{}

The order in which nodes are executed impact the lifetime of the tensors, and therefore the peak working set. Figure~\ref{fig:scheduling} illustrates a simple example where changing the operator ordering noticeably improves memory usage.

\paragraph{}

Among all possible node orderings, those prioritizing the execution of nodes that free large amounts of data while generating little output data themselves, are likely to be more efficient. However, as demonstrated in prior works~\cite{Bernstein1989OnTC, dag_schedule}, finding an optimal scheduling for a generic DAG is an NP-complete problem, which cannot be solved with a simple greedy approach.

\subsection{Optimizing Tensor Locations in Memory}
\begin{figure}
    \centering
    \includegraphics[width=0.475\textwidth]{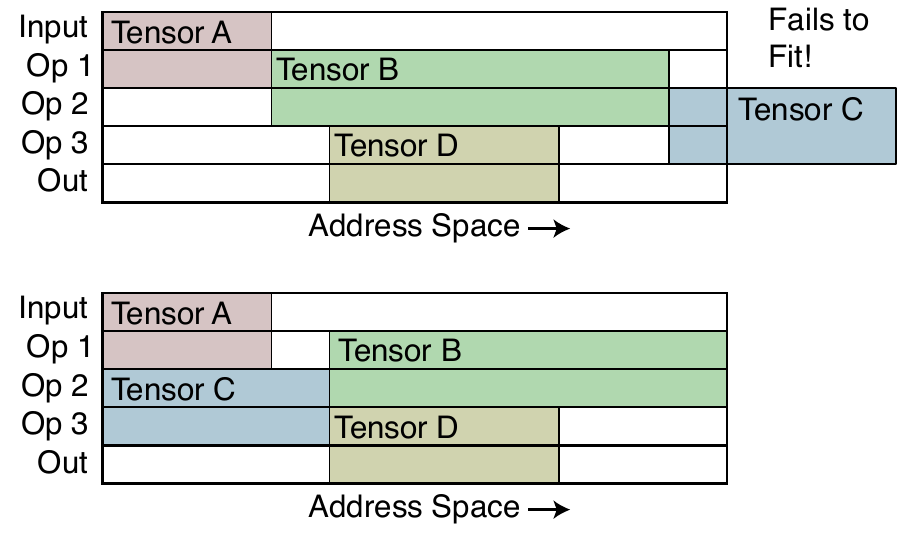}
    \caption{Memory fragmentation can cause allocation to fail. A greedy allocator (top) would not leave any room between tensor A and B, thus making the space unusable for tensor C once A is freed. \ours{} (bottom) leaves a gap between tensor A and B  to enable the reuse of the memory freed by tensor A and fits all the tensors in memory.}
    \label{fig:layoutopt}
\end{figure}

Similar to malloc-style memory allocators, the tensor allocation schemes used by typical deep learning frameworks operate online and suffers from fragmentation. Indeed, free memory is often segregated into small blocks and interspersed by memory allocated to live tensors. As a result, a significant fraction of the total memory is effectively unusable because it is divided into pieces that are not large enough to fit a tensor. Figure~\ref{fig:layoutopt} illustrates this phenomenon, and demonstrates that planning the location of each tensor ahead of time can significantly reduce the overall peak memory usage.

\section{Formulation}

We propose to take advantage of the predictability of neural network computations to proactively \underline{O}ptimize the \underline{L}ifetime and \underline {L}ocation of \underline{A}rrays (\ours{}).

\paragraph{}

We formulate the problem of optimizing the ordering of computations (which determines the tensor lifetimes) and the location of tensors in memory (which determines the amount of memory fragmentation) of generic data-flow graphs, including those used in neural network training. We encode the problem as an integer linear program, and use an off-the-shelf ILP solver to find a solution that minimizes the peak memory required to run the dataflow graph.

\paragraph{}

We solve the ILP problem ahead of time, before the training process starts. This results in a small one-time initial cost, which can be recouped over the duration of the training: as free space does not need to be found at runtime, our memory allocation and deallocation operations are much cheaper than that of a standard allocator, thus saving some time at each training step (section~\ref{practical_use}).

\subsection{DNN representation}
As mentioned in section~\ref{dag}, we model a neural network as a directed acyclic graph G = (V, E) with $n$ nodes $V = {v_1, ... , v_n}$ that represent the operators and the neural network, and $m$ edges $E = {e_1, ..., e_m}$ that encode the tensors exchanged by operators. The size in bytes of the tensor represented by edge $e_i$ is denoted as $S_i$. The source vertex of edge $e$ is denoted $src(e)$. The set of sink vertices of edge $e$ is denoted $snks(e)$.

\paragraph{}

The set of edges in the fanout of a node $v$ is denoted $fo(v)$, while the set of edges in its fanin is represented as $fi(v)$. We will also denote $fi(e)$ the set of edges in the fanin of the source vertex of $e$. We represent by $sib(e)$ the siblings to an edge $e$, that is the collection of edges that are driven by the same source vertex. 

\paragraph{}

We model time as a discrete set of timesteps $T = {t_1, ..., t_n}$. A single operation typically runs per timestep, but it is possible for several operations to execute concurrently. Therefore, we need at most $n$ timesteps to schedule a graph with $n$ operations.

\subsection{Encoding Tensor Lifetimes}

We capture the execution of a neural network as a series of tensors being allocated or preserved over time. To do this, we use two sets of binary variables:
\begin{itemize}
    \item A variable labeled $C_{e, \: t} \in \{0, 1\}$ indicates whether or not the tensor $e$ should be created (i.e. allocated) at timestep $t$ by running its source vertex.
    \item A variable named $P_{e, \: t} \in \{0, 1\}$ reflects whether tensor $e$ needs to be preserved in memory at timestep $t$ or whether it can be freed.
\end{itemize}

\paragraph{}

We leverage a set of linear constraints to ensure that the sequence of tensor creations and preservations reflects a valid execution sequence of the neural network corresponding to a feasible topological ordering of the DAG.
\paragraph{}

First, a tensor $e$ can either be created or preserved at each timestep $t$, but not both (equation~\ref{eq:live_or_preserved}). Note that it's possible for both $C_{e, \: t}$ and $C_{e, \: t}$ to be false, which indicates that the tensor does not reside in memory at this point in time.
\begin{equation}
\forall{e \in E}, \: \forall{t \in T} \quad {P_{e, \: t} + C_{e, \: t} \le 1}
\label{eq:live_or_preserved}
\end{equation}

Second, a tensor $e$ can be preserved in memory at timestep $t$ if and only if it was created or preserved at the previous timestep (equation~\ref{eq:preserve_feasibility}).

\begin{equation}
\forall{e \in E}, \: \forall{t \in T} \quad {P_{e, \: t} \le P_{e, \: t-1} + C_{e, \: t-1}}
\label{eq:preserve_feasibility}
\end{equation}

Third, to ensure that the solver does not simply avoid running any of the operators, we force every tensor $e$ to be created once through equation~\ref{eq:create_once}.
\begin{equation}
\forall{e \in E} \quad {\sum_{t \in T}{C_{e, \: t}} = 1}
\label{eq:create_once}
\end{equation}

Fourth, a tensor $e$ can only be created by running its source operator $v$. In order to do so, all the tensors in the fanin of $v$ must be present in memory (equation~\ref{eq:fanin_in_memory}).
\begin{equation}
\forall{e \in E}, \: \forall{t \in T}, \ \ \forall{f \in fi(e)} \quad {C_{e, \: t} \le P_{f, \: t}}
\label{eq:fanin_in_memory}
\end{equation}
 
Last but not least, we also need to make sure that operators with multiple outputs create their output tensors at the same time. We achieve this by tying the values of the $C_{s, t}$ variables for all the siblings $s$ to a tensor $e$ in equation~\ref{eq:multiple_outputs}.
\begin{equation}
\forall{e \in E}, \: \forall{t \in T}, \ \ \forall{s \in sib(e)} \quad {C_{s, \: t} = C_{e, \: t}}
\label{eq:multiple_outputs}
\end{equation}

The combination of constraints~\ref{eq:live_or_preserved} through~\ref{eq:multiple_outputs} ensures that all the feasible solutions to the ILP correspond to valid schedules. They guarantee that the creation timestep of each tensor corresponds to a topologically feasible ordering of the vertices of the graph. Moreover, they force the preservation in memory of each tensor from the time it is generated until the last timestep in which it is consumed. 

\subsection{Encoding Tensor Locations}
To let our solver also optimize the placement of tensors in memory, we assign an integer variable $A_e \in [0, M]$ to each tensor $e$ that encodes its base address. Here, $M=\sum_{e}{S_e}$, which corresponds to the worst case scenario where all the tensors reside concurrently in memory.

\paragraph{}

We also introduce two binary variables $a_{i,j}~\in~\{0,1\}$ and $b_{i,j}~\in~\{0,1\}$ for each pair of tensors $i$ and $j$. We constrain them through equation~\ref{eq:live} in such a way that either $a_{i,j}$ or $b_{i,j}$ is equal to 1 if both tensors reside in memory concurrently at any point in time, but can be 0 otherwise.

\begin{equation}
\begin{aligned}
   \forall{t \in T}, \: \forall{(i, j) \in E^2} \quad a_{i, j} + b_{i, j} & \le 1 \\
   a_{i, j} + b_{i, j} & \ge live_{i, t} + live_{j, t} - 1  \\
   where~live_{i, t} & = C_{i, t} + P_{i, t} \\
   and~live_{j, t} & = C_{j, t} + P_{j, t}
\end{aligned}
\label{eq:live}
\end{equation}

\paragraph{}

We use these variables to prevent the overlap of tensors that reside in memory at the same time in equations \ref{eq:below} and \ref{eq:above}.

\begin{subequations}
\begin{equation}
\forall{(i, j) \in E^2} \quad A_i + S_i - A_j \le (1 - a_{i,j}) * M
\label{eq:below}
\end{equation}
\begin{equation}
\forall{(i, j) \in E^2} \quad A_i - A_j - S_j \ge (b_{i,j} - 1) * M
\label{eq:above}
\end{equation}
\end{subequations}

If $a_{i,j}$ takes the value 1, equation~\ref{eq:below} degenerates into $A_i~+~S_i~\le~A_j$. This forces tensor $i$ to reside below tensor $j$ in memory. Similarly, equation~\ref{eq:above} degenerates into $A_i~\ge~A_j~+~S_j$ when $b_{i,j}$ takes the value 1, which forces tensor $i$ to be placed above tensor $j$. On the other hand, if $a_{i,j}$ and $b_{i,j}$ take the value 0, equations~\ref{eq:below} and ~\ref{eq:above} hold for any value of $A_i$ and $A_j$ in the range $[0, M]$. In other words, they don't impose further restrictions on the location of $e_i$ and $e_j$.

\paragraph{}

Put altogether, constraints~\ref{eq:live}, ~\ref{eq:below}, and~\ref{eq:above} ensure that tensors can share the same memory space if and only if their lifetimes do not overlap.

\subsection{Minimizing Peak Memory Usage}
We track the peak memory usage by introducing a variable $peak\_mem$ that we constrain as follow:
\begin{equation}
\forall{e \in E} \quad {A_e + S_e \le peak\_mem}
\label{eq:peak_address}
\end{equation}

We find the schedule of operators and memory location of tensors that optimizes the memory usage of the neural network by feeding  program~\ref{eq:complete_ilp} to an ILP solver.

\begin{equation}
\begin{aligned}
& \arg \min_{C, P, A}{peak\_mem} \\
subject~to~&(\ref{eq:live_or_preserved}), (\ref{eq:preserve_feasibility}), (\ref{eq:create_once}), (\ref{eq:fanin_in_memory}), (\ref{eq:multiple_outputs}), \\ &(\ref{eq:live}), (\ref{eq:below}), (\ref{eq:above}), (\ref{eq:peak_address})
\end{aligned}
\label{eq:complete_ilp}
\end{equation}

\subsection{Decoding the ILP Result}
\label{decoding}
Given a feasible solution to our ILP, we generate an optimized execution sequence of operations $ES$ = ($s_1$, ..., $s_k$) for the neural network using function~\ref{alg:execution_plan}.

\paragraph{}

\begin{algorithm}
\begin{algorithmic}
\STATE \COMMENT{Converts the output of the ILP into an optimized}
\STATE \COMMENT{execution sequence of operations $seq$.}
\STATE seq = []
\FOR{$t$ {\bfseries in} $T$}
  \FOR{$e$ {\bfseries in} $E$}
    \IF{$C_{e,t}$ = 1}
      \STATE add \texttt{execute(src(e))} to seq
    \ENDIF
  \ENDFOR
\ENDFOR
\RETURN seq
\end{algorithmic}
\caption{GenerateExecutionSequence($C$)}
\label{alg:execution_plan}

\end{algorithm}

Note that our algorithm may generate duplicate $execute(v)$ statements in the case where a node $v$ has multiple edges in its fanout. These redundant statements need to be removed to ensure the correctness of the final program.

\paragraph{}

Tensors are stored in a shared preallocated buffer $B$ sized to accommodate the peak memory usage. The value of each $A_e$ variable represents the offset location of tensor $e$ in $B$.

\paragraph{}

We can map memory allocation requests to addresses over multiple iterations of the training loop as follow. We'll assume that each operator generates a single output tensor for the sake of simplicity, but our approach generalizes to handle operators with multiple outputs. The $k_{th}$ memory allocation request corresponds to the tensor generated by the operator located at position $k\bmod{|V|}$ in the execution sequence $ES$. This tensor $e$ is to be located at address $A_B+A_e$, where $A_B$ is the base address of buffer $B$. Memory deallocation requests are no-ops.

\section{Scaling to Large Neural Networks}

Our formulation requires $2\times|E|\times|V|$ binary variables since we have one $C$ and one $P$ variable per tensor per timestep, as well as $|A|$ integer variables to track tensor addresses. Additionally, we create $\bigO(|V|\times|E|)$ constraints to encode tensor precedence and life cycle requirements, and $\bigO(|V|\times|E|^2)$ constraints to ensure that tensors never overlap in memory.

\paragraph{}

We develop five techniques to reduce the complexity of the ILP formulation and enable our approach to scale well. This permits \ours{} to optimize the memory usage of neural networks with complex tensor computation graphs comprised of thousands of vertices and edges.

\subsection{Bounding Lifetime Ranges}

All of the input tensors of a node must reside in memory for it to run at a given timestep. This means that all the operators in the immediate fanin of the node must have been run at least one timestep prior. As a result, we can identify the earliest timestep $ASAP(v)$ (``as soon as possible'') during which a node $v$ can run. $ASAP(v)$ is the longest distance from $v$ to an input of the neural network, which is computed in linear time using a simple depth first search traversal of the graph~\cite{asap_alap}. Using the same approach, we can also identify the latest timestep $ALAP(v)$ (``as late as possible'') at which a node $v$ can run, which is the longest distance from $v$ to an output of the neural network.

\paragraph{}

A node $v$ can only run within the span $[ASAP(v), ALAP(v)]$. Since tensors are created when their source node is run, a variable $C_{e, t}$ will always be false outside the span of their source node (Equation~\ref{eq:span}).
\begin{equation}
\begin{gathered}
    SPAN(v)=[ASAP(v), ALAP(v)] \\
    \forall{e \in E}, \:\: \forall{t \notin{SPAN(src(e))}} \quad C_{e, t} = 0
\end{gathered}
\label{eq:span}
\end{equation}

Furthermore, a tensor only needs to be preserved in memory until all its sink operators have run. This enables us to define the Maximum Useful Lifetime (MUL) range of a tensor, and set the variable $P_{e, t}$ for a tensor $e$ to false outside of this range (Equation~\ref{eq:mul}).

\begin{equation}
\begin{gathered}
    MUL(e)=[ASAP(src(e)), \underset{s \in snks(e)}{max}{ALAP(s)}] \\
    \forall{e \in E}, \: \: \forall{t \notin{MUL(e)}} \quad P_{e, t} = 0
\end{gathered}
\label{eq:mul}
\end{equation}

Additionally, tensors must be preserved in memory from the time they are created until their last sink node has run. Therefore, $P_{e, t}$ must always be true from the last timestep at which $e$ can be created until the earliest timestep at which its last sink can run (Equation~\ref{eq:pres}). 

\begin{equation}
\begin{gathered}
    PRES(e)=[ALAP(src(e)+1, \underset{s \in snks(e)}{max}{ASAP(s)}] \\
    \forall{e \in E}, \: \: \forall{t \in{PRES(e)}} \quad P_{e, t} = 1
\end{gathered}
\label{eq:pres}
\end{equation}

This enables us to reduce the number of timesteps to track for each tensor. In the best case scenario, where a neural network is a linear sequence of operators, the span of each node $v$ is reduced to a single timestep, and we can derive the values of all the $C_{e, t}$ and $P_{e, t}$ purely from the structure of the graph. However, in the opposite extreme case where a neural network consists exclusively of operators that can run in parallel, we cannot infer any of the values of the $C_{e, t}$ and $P_{e, t}$ variable. The structure of real neural networks lies somewhere between these two extremes.

\subsection{Leveraging Precedence Constraints}
We simplify our memory placement formulation by skipping constraints \ref{eq:live} \ref{eq:below} and \ref{eq:above} whenever we can determine that two tensors can never reside in memory at the same time. We exploit two sufficient conditions to achieve this.

\paragraph{}

First, we leverage the Maximum Useful Lifetime ranges from our ASAP/ALAP analysis. If the MUL ranges of two tensors do not overlap, they will never be present concurrently in memory.

\paragraph{}

\begin{figure}
    \centering
    \includegraphics[width=0.48\textwidth]{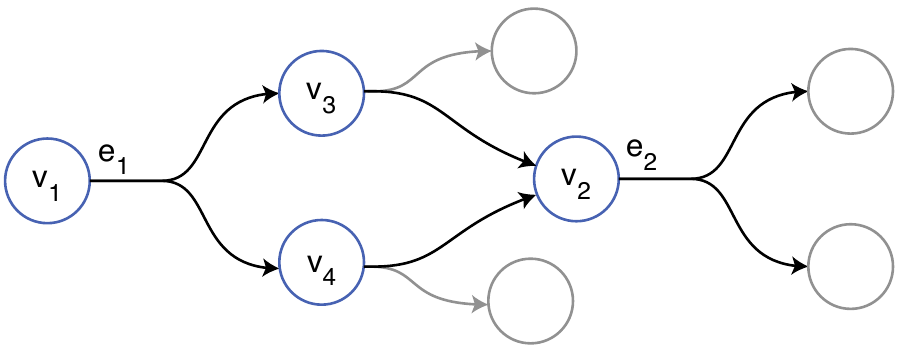}
    \caption{Edge precedence: $e_1\prec_{prec}e_2$ since the sinks $v_3$ and $v_4$ of $e_1$ are both in the transitive fanin of the source node of $e_2$, and $e_1$ and $e_2$ have no vertex in common.}
    \label{fig:edge_precendence}
\end{figure}

We complement this first condition with a precedence analysis. If a vertex $v_2$ is reachable from another vertex $v_1$ (i.e. if $v_1$ is in the transitive fanin of $v_2$), the corresponding operator $v_1$ must be run before operator $v_2$. Therefore, if all the sink vertices of an edge $e_1$ are in the transitive fanin of the source vertex of an edge $e_2$, $e_1$ and $e_2$ can only be present in memory if there is a vertex $v$ such that $e_1$ is one of the fanout edges of $v$ and $e_2$ is one of its fanin edges (Figure~\ref{fig:edge_precendence}). We call this condition $\prec_{prec}$, and if either condition $e_1\prec_{prec}e_2$ or $e_2\prec_{prec}e_1$ holds $e_1$ and $e_2$ can never reside together in memory.

\paragraph{}
We use a simple depth-first search (Function~\ref{alg:transitive_fanin}) to determine whether a vertex $v_2$ is reachable from a vertex $v_1$. We leverage memoization to ensure that answering the query for a pair $(v_1, v_2)$ yields to constant time queries for all future queries $(v, v_2)$ that involve a vertex $v$ on a path from $v_1$ to $v_2$.  

\begin{algorithm}
\begin{algorithmic}

\STATE \COMMENT{Returns true iff v2 can be reached from v1.}
\IF{$(v1, v2)$ {\bfseries in} $cache$}
    \RETURN $cache[(v1, v2)]$
\ENDIF
\FOR{$f$ {\bfseries in} $fi(v2)$}
    \IF{$src(f) = v1$}
        \STATE $cache[(v1, v2)] \gets \TRUE$
        \RETURN \TRUE 
    \ENDIF
    \IF{$IsInTransitiveFanin(v1, src(f))$}
        \STATE $cache[(v1, v2)] \gets \TRUE$
        \RETURN \TRUE 
    \ENDIF
\ENDFOR
\STATE $cache[(v1, v2)] \gets \FALSE$
\RETURN \FALSE

\end{algorithmic}
\caption{IsInTransitiveFanin(v1, v2, cache)}
\label{alg:transitive_fanin}
\end{algorithm}

\subsection{Enforcing Early Memory Deallocations}
Running a weight update operation enables the freeing of the corresponding gradient tensor. Applying these updates early reduces memory pressure, and conversely, there is no benefit to delaying their execution. To speed up the tensor lifetime optimization process, we prevent the solver from considering running these nodes late in the computation.

\paragraph{}

To achieve this, we look for a good anchor node, and add an edge of size 0 (which we call a control edge) from the gradient update node to this anchor node. The control edge forces the ALAP time of the gradient node to be less than that of the anchor node, but has no impact on memory usage since its size is 0. The anchor node must meet two criteria:
\begin{itemize}
    \item Its level must be greater than that of the weight update node in the graph levelization~\cite{levelization}. This prevents the introduction of a loop through the control edge in the graph. 
    \item A good anchor candidate must be scheduled early in the computation itself. We determine this by running the levelization on a copy of the DAG in which the directions of all the edges are reversed, and looking for a node with a high backward level.  
\end{itemize}

\paragraph{}

We start the search for anchor nodes in the immediate fanin of the weight update vertex, and progressively expands the search radius until we find a suitable node as detailed in Functions~\ref{alg:gradient_update} and~\ref{alg:cand_search}.

\begin{algorithm}
\begin{algorithmic}
   \STATE \COMMENT{Adds control edges to G to force the weight update nodes}
   \STATE \COMMENT{to run early. The pseudo code for $FindCandidate$ is}
   \STATE \COMMENT{provided in Function~\ref{alg:cand_search}.}

   \STATE $fwd\_lvl \gets ComputeLevelization(G)$
   \STATE $bwd\_lvl \gets ComputeReverseLevelization(G)$
   
   \FOR{$v$ {\bfseries in} gradient update nodes of $G$}
      \STATE $min\_fwd\_level \gets fwd\_lvl[v]$
      \STATE $best\_bwd\_level \gets -1$
      \STATE $best\_anchor \gets none$
      \STATE $search\_starts \gets set(v)$
      \STATE $visited \gets hashtable()$
      \WHILE{\NOT $best\_anchor$ \AND $len(search\_starts) > 0$}
        
        \STATE $next\_starts \gets set()$
        \FOR {$v$ {\bfseries in} $search\_starts$}
          \FOR {$f$ {\bfseries in} $fi(v)$}
            \STATE add $src(f)$ to $next\_starts$
          \ENDFOR
        \ENDFOR
        
        \STATE $search\_starts \gets next\_starts$
        
        \FOR {$src$ {\bfseries in} $search\_starts$}
            \STATE $candidate, level \gets FindCandidate(src, $
            \STATE $\quad fwd\_lvl, bwd\_lvl, min\_fwd\_level, visited)$
            \IF {level > $best\_bwd\_level$}
                \STATE $best\_bwd\_level \gets level$
                \STATE $best\_anchor \gets candidate$
            \ENDIF
                    
        \ENDFOR
        \IF{$best\_anchor$}
            \STATE add control edge from $v$ to  $best\_anchor$
        \ENDIF
      \ENDWHILE
    \ENDFOR
\end{algorithmic}
\caption{EnforceEarlyWeightUpdates(G)}
\label{alg:gradient_update}
\end{algorithm}

\begin{algorithm}
\begin{algorithmic}
   \STATE \COMMENT{Find a candidate anchor node starting the search from}
   \STATE \COMMENT{node v.} 

    \IF{v {\bfseries in} $visited$}
        \RETURN $visited[v]$
    \ENDIF
        
    \STATE $best\_bwd\_level \gets -1$
    \STATE $best\_candidate \gets none$

    \FOR{$f$ {\bfseries in} $fo(v)$}
        \FOR{$snk$ {\bfseries in} $snks(f)$}
            \IF{$bwd\_lvl[snk] \le best\_bwd\_level$}
                \STATE {\bfseries continue}
            \ENDIF
            \IF{$fwd\_lvl[snk] \le min\_fwd\_level$}
                \STATE $candidate, level \gets FindCandidate(snk, $
                \STATE $\quad fwd\_lvl, bwd\_lvl, min\_fwd\_level, visited)$
                \IF{$level > best\_bwd\_lvl$}
                    \STATE $best\_bwd\_level \gets level$
                    \STATE $best\_candidate \gets candidate$
                \ENDIF
            \ELSE
                \STATE $best\_bwd\_level \gets bwd\_lvl[snk]$
                \STATE $best\_candidate \gets snk$
            \ENDIF
        \ENDFOR
    \ENDFOR

    \STATE $visited[v] \gets (best\_candidate, best\_bwd\_level)$
    \RETURN $(best\_candidate, best\_bwd\_level)$
        
\end{algorithmic}
\caption{FindCandidate(v, fwd\_lvl, bwd\_lvl, min\_fwd\_lvl, visited)}
\label{alg:cand_search}
\end{algorithm}

\subsection{Splitting the Problem}
We noticed empirically that \ours{} is always able to fully eliminate memory fragmentation. Therefore, without loosing optimality, we can divide the problem in two subproblems that are faster to solve sequentially.

\paragraph{}

We first look for the schedule of operations that leads to the smallest peak memory usage under the assumption that we will be able to locate tensors in memory later on without introducing any fragmentation. We compute this peak memory usage metric, $peak\_mem\_no\_frag$, by leveraging the $C$ and $P$ variables in equation~\ref{eq:peak_mem}:
\begin{equation}
\forall{t \in T} \quad {\sum_{e \in E}{(C_{e, \: t}+ P_{e, \: t}) \times S_e} \le peak\_mem\_no\_frag}
\label{eq:peak_mem}
\end{equation}

The problem of optimizing the lifetime of tensors without considering fragmentation is then formulated in equation~\ref{eq:ilp_for_scheduling}:

\begin{equation}
\begin{aligned}
 \arg \min_{C, P}&\:{peak\_mem\_no\_frag} \\
subject~to&~(\ref{eq:live_or_preserved}), (\ref{eq:preserve_feasibility}), (\ref{eq:create_once}), (\ref{eq:fanin_in_memory}), (\ref{eq:multiple_outputs}), (\ref{eq:peak_mem})
\end{aligned}
\label{eq:ilp_for_scheduling}
\end{equation}

\paragraph{}

Given a solution to~(\ref{eq:ilp_for_scheduling}) we tackle the problem of optimizing the location of tensors in memory by solving equation~\ref{eq:ilp_for_addresses}:

\begin{equation}
\begin{aligned}
& \arg \min_{A}{peak\_mem} \\
subj&ect~to~(\ref{eq:live}), (\ref{eq:below}), (\ref{eq:above}), (\ref{eq:peak_address})
\end{aligned}
\label{eq:ilp_for_addresses}
\end{equation}

\subsection{Finding Suitable Memory Locations for Activations}

\begin{figure}
    \centering
    \includegraphics[width=0.48\textwidth]{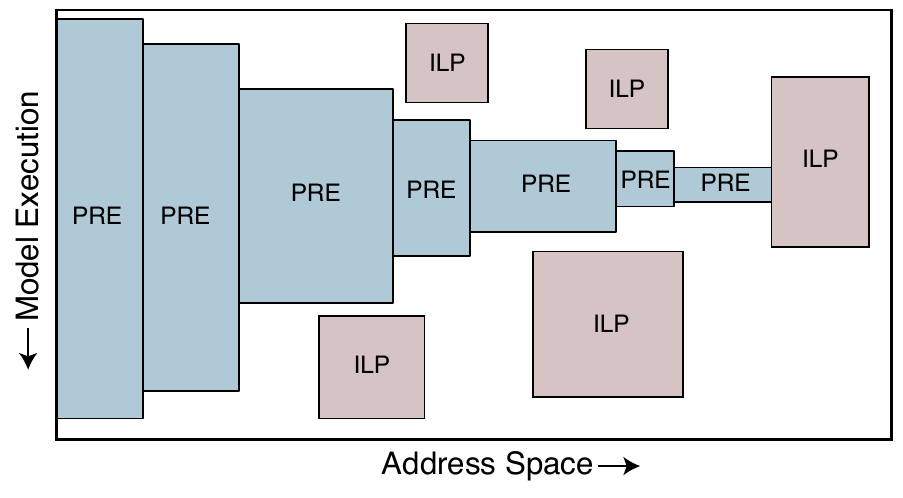}
    \caption{Mixed tensor placement: function~\ref{alg:mem_placement} assigns addresses to the tensors marked PRE, while the ILP solver assigns addresses to the tensors marked ILP.}
    \label{fig:address_generation_opt}
\end{figure}

DNN gradients are computed in reverse order of the activations. Since an activation is preserved in memory until the corresponding gradient computation takes place, the earlier an activation tensor is allocated the later it is freed. We place these tensors in memory in a manner that maximizes the usability of the unallocated memory. After determining the tensor lifetimes with equation~\ref{eq:ilp_for_scheduling}, we order them by decreasing SPAN (equation~\ref{eq:span}), and place them at increasing memory addresses to form what looks like a pyramid as illustrated in Figure~\ref{fig:address_generation_opt}. This process is detailed in algorithm~\ref{alg:mem_placement}. The ILP solver assigns addresses to the remaining tensors.

\paragraph{}

This speeds up the ILP solver in two ways. First, the number of tensors it needs to place in memory is decreased significantly. Second, the address space available for these tensors is noticeably reduced.

\paragraph{}

Although it is a heuristic, this memory preplacement approach does not compromise \ours{}'s ability to eliminate fragmentation as we will show in section~\ref{fragmentation_results}.

\begin{algorithm}
\begin{algorithmic}
    \STATE \COMMENT{Find memory locations for a subset of the tensors.}
    \STATE $min\_start \gets 0$
    \STATE $max\_end \gets \infty$
    \STATE $base\_address \gets 0$
    \STATE $processed \gets set()$
    \WHILE{$max\_end$ > $min\_start$}
        \STATE $max\_duration \gets 0$
        \STATE $next\_step \gets none$
            \FOR{$e$ {\bfseries in} $E$}
            \STATE $first\_use \gets ASAP(src(e))$
            \STATE $last\_use \gets \max_{s \in snks(e)}{ALAP(s)}$

            \IF{$first\_use$ < $min\_start$}
                \STATE {\bfseries continue}
            \ENDIF
            \IF{$last\_use$ > $max\_end$}
                \STATE {\bfseries continue}
            \ENDIF
            \IF{$tensor$ {\bfseries in} $processed$}
                 \STATE {\bfseries continue}
            \ENDIF
            \STATE $duration \gets last\_use$ - $first\_use$
            \IF{$duration$ > $max\_duration$}
                \STATE $max\_duration \gets duration$
                \STATE $next\_step \gets tensor$
            \ENDIF
            \ENDFOR
            \IF{{\bfseries not} $next\_step$}
                \STATE {\bfseries break}
            \ENDIF
            \STATE $A_{next\_step} \gets base\_address$
            \STATE $base\_address \gets base\_address + S_{next\_step}$
            \STATE $min\_start \gets first\_use$
            \STATE $max\_end \gets last\_use$
            \STATE add $next\_step$ to $processed$
    \ENDWHILE
    
\end{algorithmic}
\caption{PreAllocateAddresses(G)}
\label{alg:mem_placement}
\end{algorithm}

\section{Experiments}

We measured the impact of \ours{} on the memory usage of DNN training. We tried to answer the following questions:
\begin{itemize}
    \item How effective are our two strategies of node reordering and address generation at reducing memory usage?
    \item How applicable is our approach? Can one reasonably expect to benefit from it on their use case, or is it more effective in some scenarios than others?
    \item How practical are our algorithms? Can they be applied to large neural networks in a reasonable amount of time?
\end{itemize}

\subsection{Experimental Setup}
We implemented \ours{} on top of PyTorch version 1.11~\cite{PyTorch} with  torchtext 0.12 and torchvision 0.12. We leveraged torch.FX to convert neural networks into executable sequences of operator calls, and reconstructed the computation graphs from the operator arguments. We encoded and solved the memory optimizations problems~(\ref{eq:complete_ilp}), (\ref{eq:ilp_for_scheduling}) and (\ref{eq:ilp_for_addresses}) using Gurobi version 9.1.1~\cite{Gurobi}. We translated the Gurobi results into optimized execution sequences and memory locations as described in section~\ref{decoding}.

We ran all our experiments on a workstation featuring a Intel Xeon Gold 6138 CPU running at 2.0 GHz and a NVidia A100 GPU with 40GB of memory. 

\subsection{Methodology}
We evaluated \ours{} on a comprehensive set of neural networks. We included the ResNet~\cite{resnet} and Transformer~\cite{transformer} models since they are ubiquitous and used in many downstream tasks: the former introduced the concept of residual connection, and the later popularized the attention mechanism. We also included neural networks designed for specific tasks, such as computer vision (AlexNet~\cite{alexnet}, VGG~\cite{vgg}, GoogleNet~\cite{googlenet}), video understanding (ResNet3D~\cite{resnet3d}), and language modelling (XLM-R~\cite{xlm-r}).

\paragraph{}

In addition to these models that were designed to run on datacenter hardware, we also evaluated our approach on EfficientNet~\cite{efficientnet} and MobileNet~\cite{mobilenet}. These two neural networks were tailored to run in resource constrained environments such as edge devices. Additionally, we trained the neural networks at batch size 1 and 32. Batch size 1 is commonly used when training a model on devices with limited memory capacity, while batch size 32 is used often when running in datacenters.

\paragraph{}

To be representative of the evolution of DNN designs over time, we made sure our models cover almost a decade of machine learning research, starting with AlexNet~\cite{alexnet} which was published back in 2012 and ending with VIT~\cite{vit} which was released in 2020. We also tested our approach on MNASNet~\cite{mnasnet}, a model designed by a computer using an automated process called neural architecture search~\cite{nas}.

\paragraph{}

To validate the scalability of our solution, we tested it on neural networks as small as Alexnet~\cite{alexnet} (118 operators) and as large as XLM-R~\cite{xlm-r} (2007 operators).

\subsection{Memory Reduction Resulting from Node Reordering}

To evaluate the impact of our tensor lifetime optimization, we compared the peak memory necessary to run various neural networks when using the PyTorch node ordering and the node ordering determined by algorithm~\ref{eq:ilp_for_scheduling}. In our measurements, we eliminated the impact of memory fragmentation by recording the peak memory PyTorch operators need to request to run these models under both node orderings instead of the actual memory usage.

\begin{figure}
    \centering
    \includegraphics[width=0.48\textwidth]{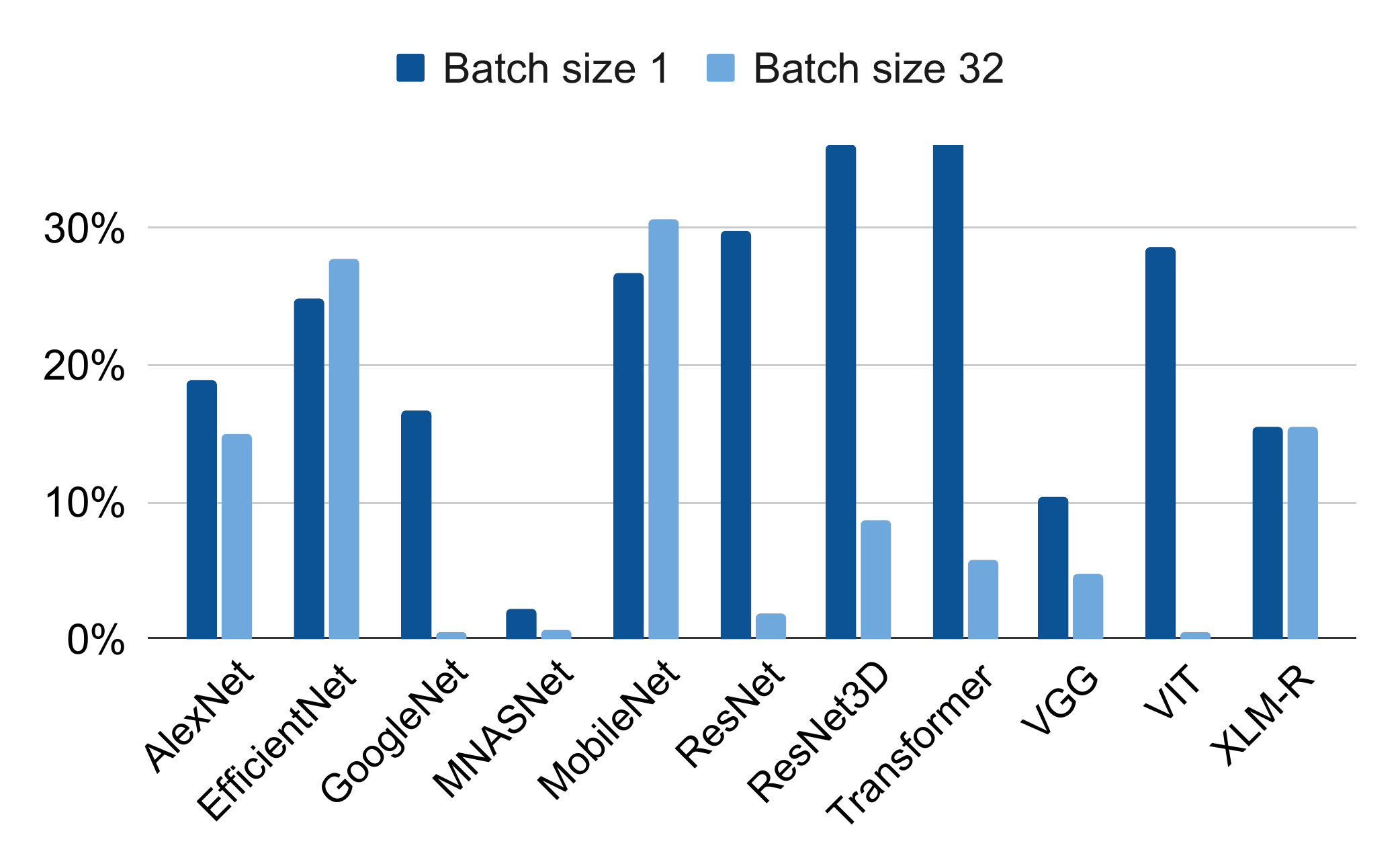}
    \caption{Peak memory reduction (in \%) compared to PyTorch as a result of our node reordering.}
    \label{fig:node_ordering}
\end{figure}

\paragraph{}

We find that \ours{} reduces peak memory usage by up to 38\% compared to PyTorch (Figure~\ref{fig:node_ordering}). On average, our solution achieves a reduction of 22.5\% at batch size 1 and 10.1\% at batch size 32.

\paragraph{}

The activations generated during the forward pass are preserved in memory for the backward pass of the training. As a result, \ours{} has little to no ability to decrease the memory usage of the forward pass. On the other hand, the order of the computation and application of the gradients with respect to the weights offers a great deal of flexibility, which \ours{} leverages to decrease the memory usage of the backward pass. However, the gradients are are roughly smaller than the activations by a factor of batch size. Therefore, at larger batch sizes, most of the memory is used to store activations, while at smaller batch size gradients represent a larger fraction of the total. This explains why our approach is more effective at small batch sizes.

\subsection{Memory Savings Coming from Address Generation}
\label{fragmentation_results}
We define the fragmentation of a memory allocator as the difference between the memory the allocator needs to reserve from the hardware $MR$ and the size of the resident set $RS$. We measure it when $MR$ reaches its peak value using the ratio $(MR-RS)/MR$. 

\paragraph{}

\begin{figure}
    \centering
    \includegraphics[width=0.48\textwidth]{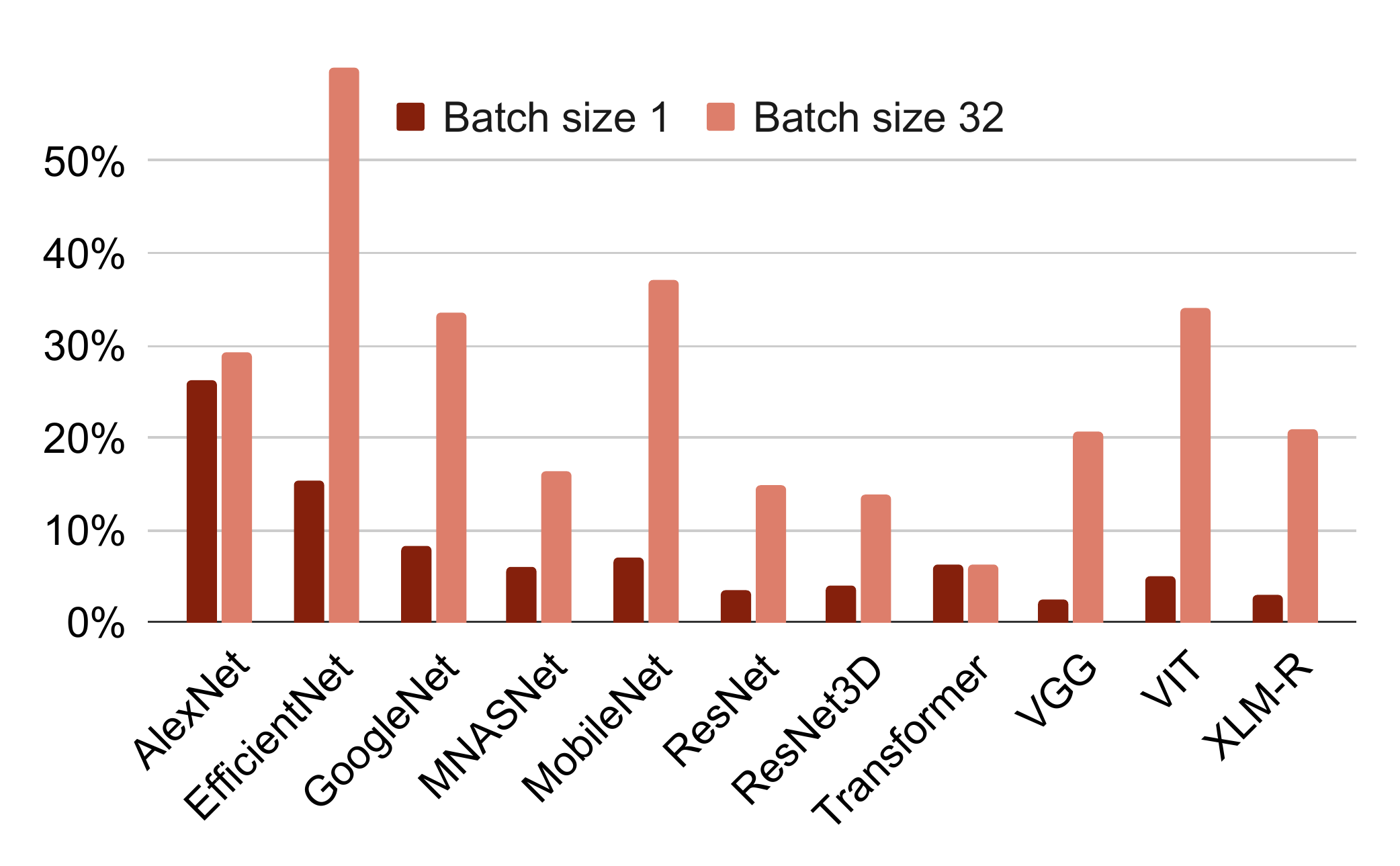}
    \caption{PyTorch memory fragmentation (in \%) during training at various batch sizes. Our method fully eliminates fragmentation.}
    \label{fig:training_frag}
\end{figure}

In all scenarios our address generator can completely eliminate memory fragmentation. By contrast, PyTorch suffered from an average fragmentation of 7.9\% at batch size 1, and 26.1\% at batch size 32 (Figure~\ref{fig:training_frag}). The PyTorch memory allocator uses a different strategy for small and large objects, which could explain why fragmentation is significantly worse for the larger batch size. However it is unclear whether it could be modified to better handle large tensors without introducing other drawbacks.

\subsection{Node Ordering Time}
\begin{figure}
    \centering
    \includegraphics[width=0.48\textwidth]{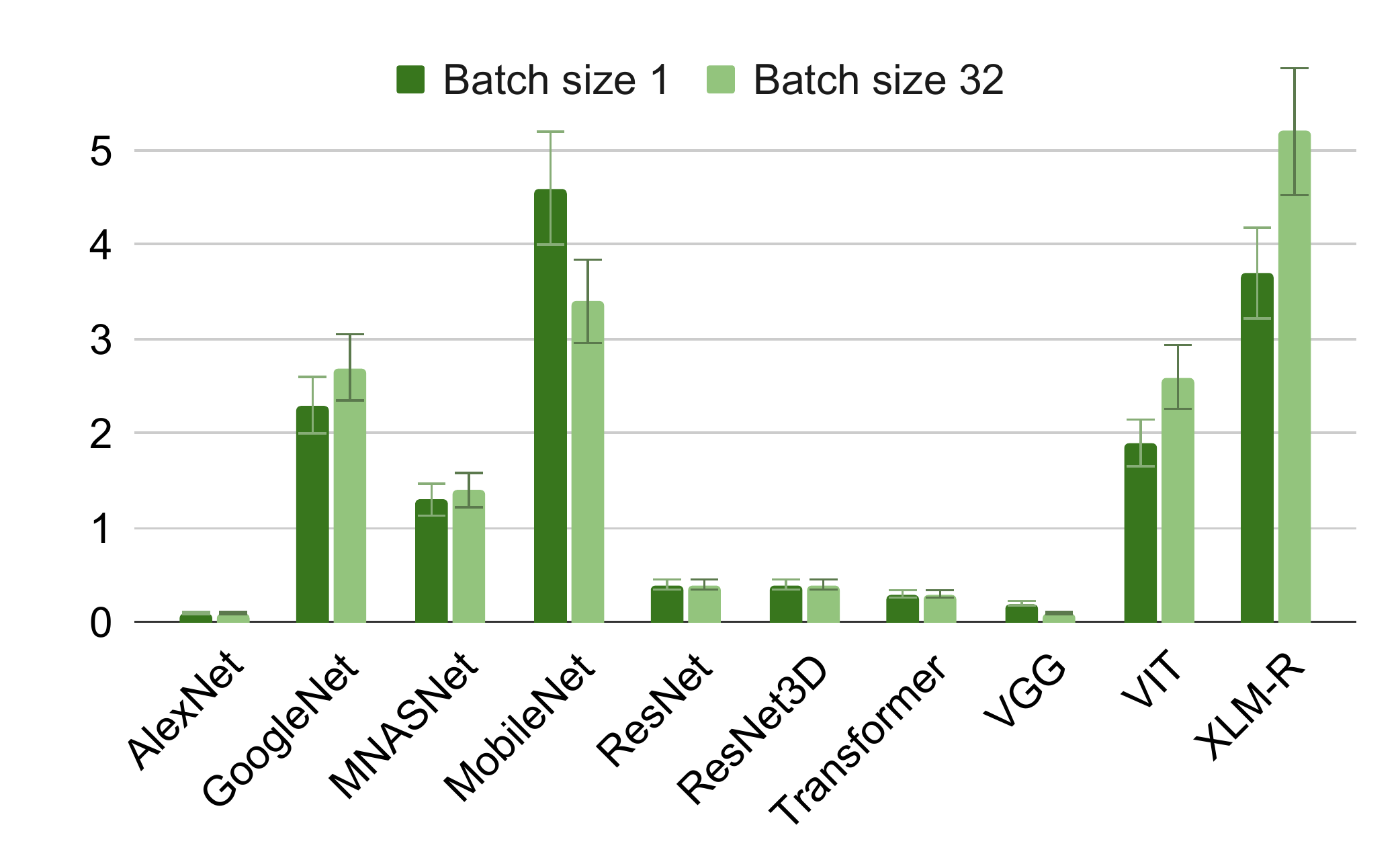}
    \caption{Node ordering times (in seconds) for  training graphs at batch sizes 1 and 32. We tack the  result for EfficientNet in figure~\ref{fig:node_ordering_over_time} to improve readability.}
    \label{fig:node_ordering_time}
\end{figure}

\begin{figure}
    \centering
    \includegraphics[width=0.48\textwidth]{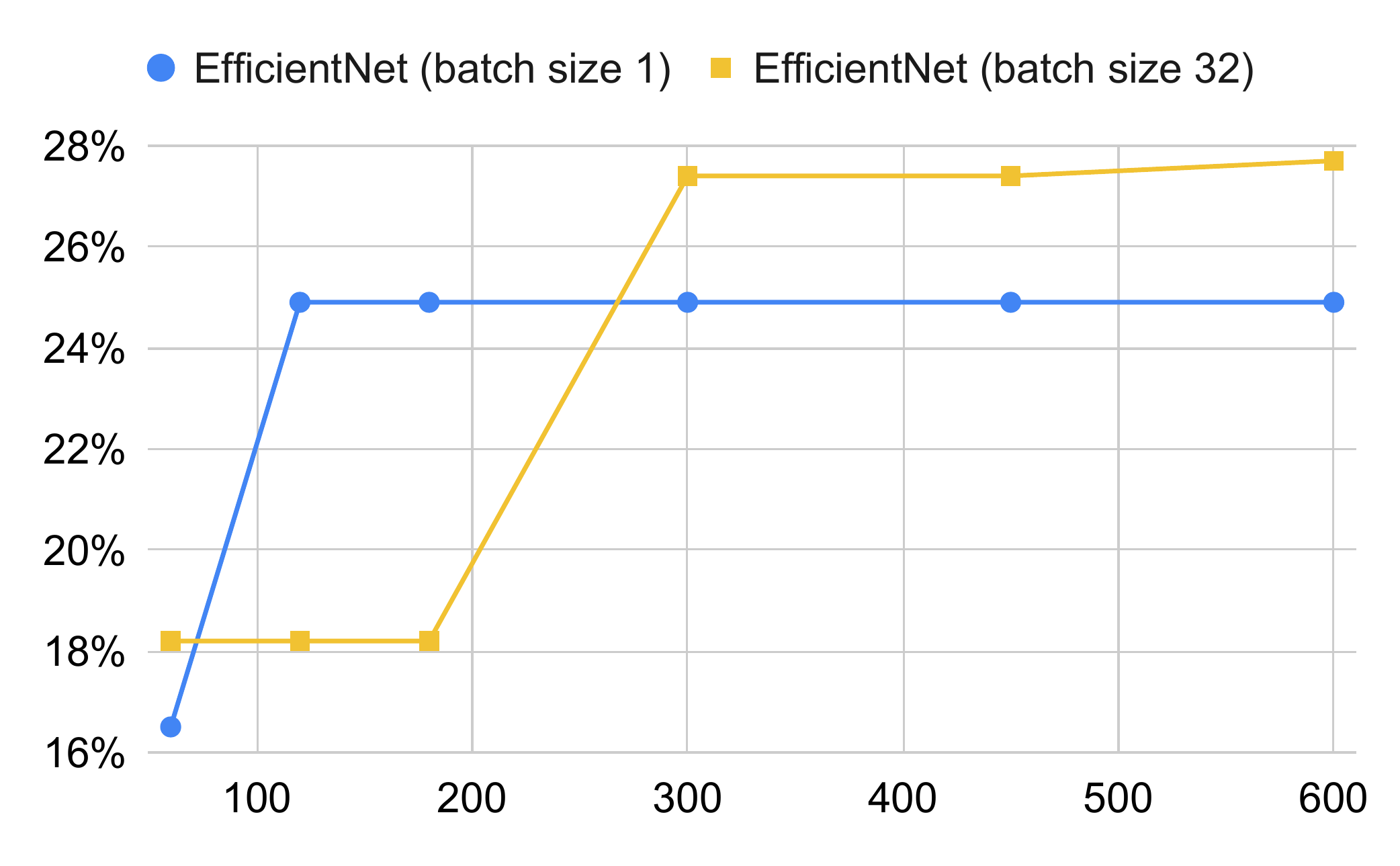}
    \caption{Memory saved (in \%) as the results of node reordering over time (in seconds). \ours{} finds the optimal solution given enough time (10 minutes).}
    \label{fig:node_ordering_over_time}
\end{figure}

Solving for equation~\ref{eq:ilp_for_scheduling} to optimize node orderings takes a median of $1.4\pm0.2$ seconds. Excluding the outlier EfficientNet, the worst case optimization time is 5.2 seconds, and the best case is 100 milliseconds (Figure~\ref{fig:node_ordering_time}). 
For EfficientNet, \ours{} needs 2 minutes to find the optimal node ordering at batch size 1 and 5 minutes to find a solution that is within 1\% of optimal at batch size 32 (Figure~\ref{fig:node_ordering_over_time}).

\paragraph{}
\ours{} only needs a small fraction of the time it takes to train a neural network to optimize the lifetime of tensors and significantly reduce the peak memory usage. 

\subsection{Address Generation Time}
\begin{figure}
    \centering
    \includegraphics[width=0.48\textwidth]{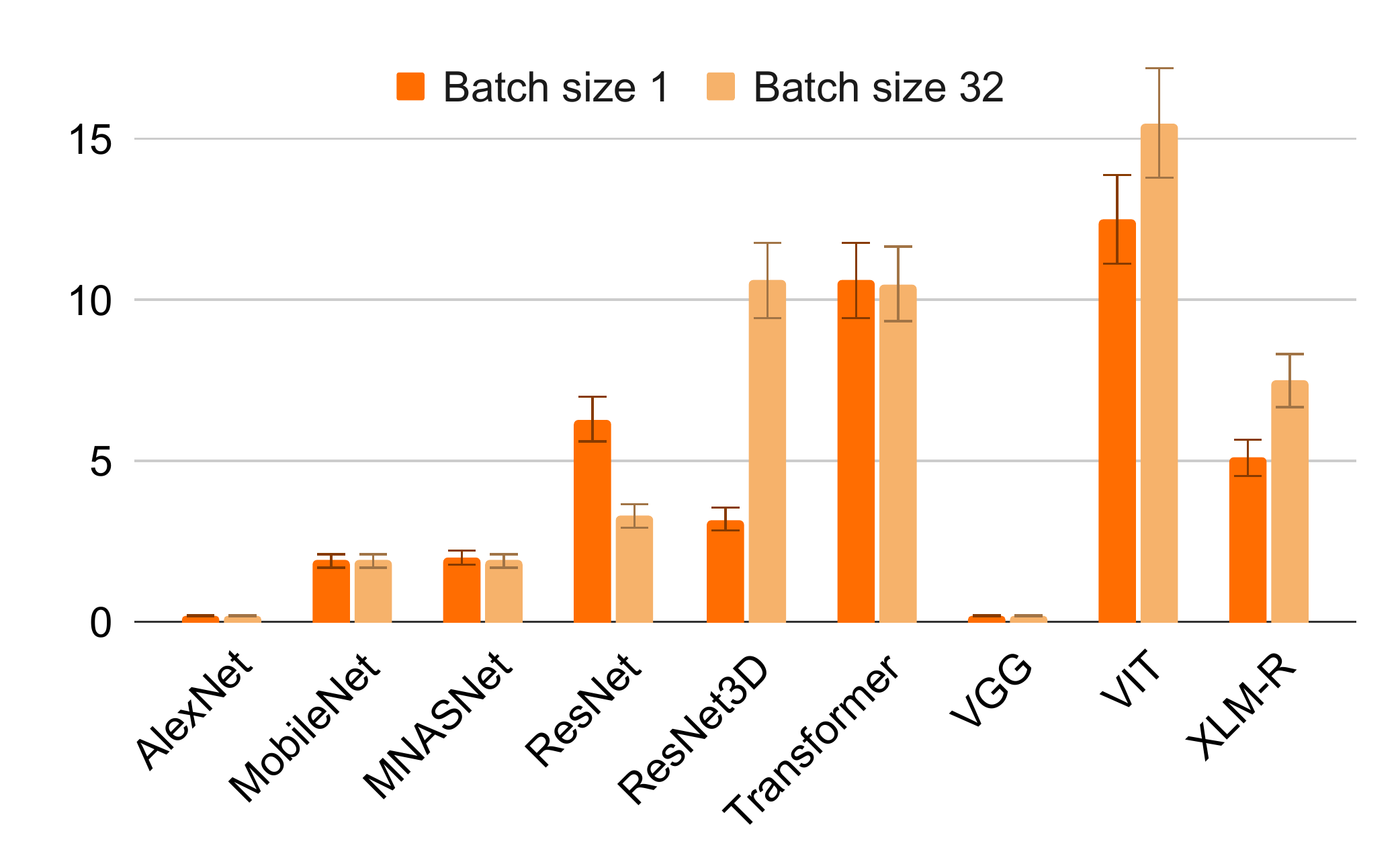}
    \caption{Fragmentation elimination times (in seconds) for  training graphs at batch sizes 1 and 32. The optimization times for EfficientNet and GoogleNet are tracked separately in figure~\ref{fig:fragmentation_over_time}.}
    \label{fig:fragmentation_time}
\end{figure}

\begin{figure}
    \centering
    \includegraphics[width=0.48\textwidth]{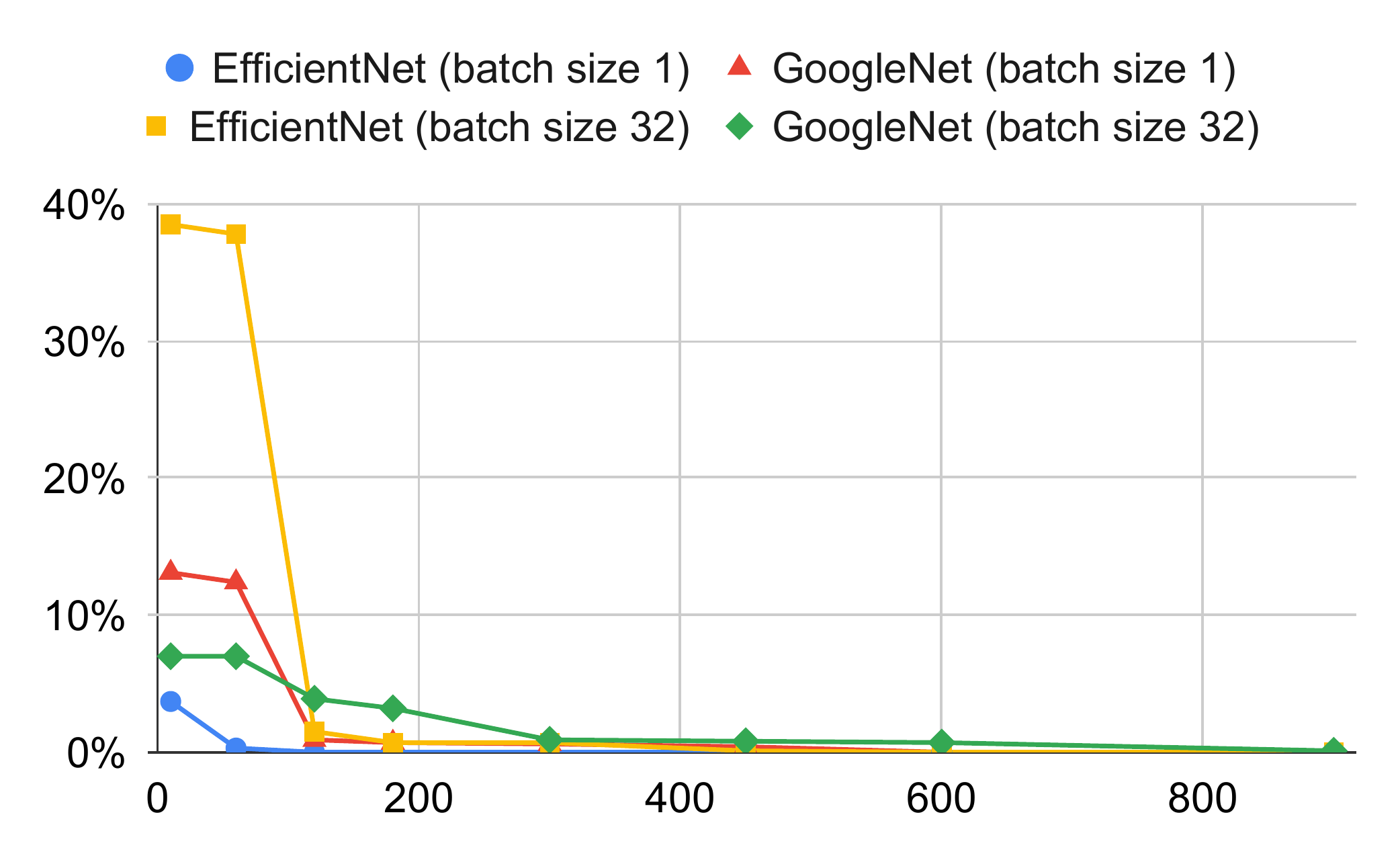}
    \caption{Memory fragmentation as a percentage of total memory usage for various optimization times (in seconds). The memory fragmentation decreases quickly towards 0 as the solver is given more time to find an optimal solution.}
    \label{fig:fragmentation_over_time}
\end{figure}

Leveraging equation~\ref{eq:ilp_for_addresses}, \ours{} eliminates fragmentation in a median time of $5.7\pm0.6$ seconds (Figure~\ref{fig:fragmentation_time}). While it takes significant effort to find optimal solutions for GoogleNet and EfficientNet, \ours{} reduced fragmentation to less than 1\% in 5 minutes or less for these two models. We plot the evolution of the memory fragmentation over time in these two cases in Figure~\ref{fig:fragmentation_over_time}.

\subsection{Practical Use}
\label{practical_use}
\begin{figure}
    \centering
    \includegraphics[width=0.48\textwidth]{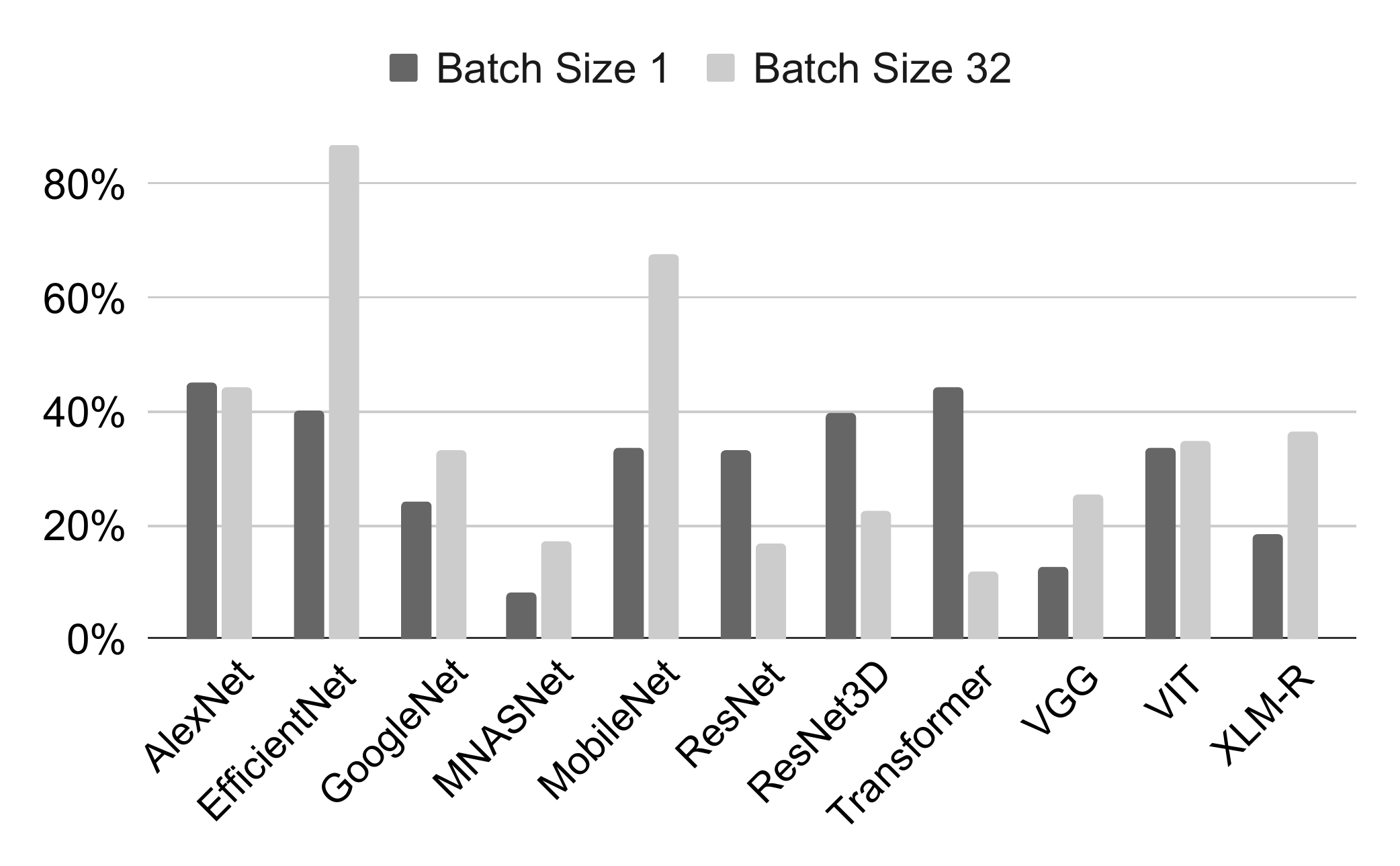}
    \caption{Total reduction in peak memory usage (in \%) during training at various batch sizes compared to PyTorch. The optimization process is capped at 10 minutes.}
    \label{fig:total_memory_savings}
\end{figure}

Since our optimizer runs ahead of time, it's important that it completes its task in a short amount of time to avoid negatively impacting the user friendliness of the overall system. We achieved a balance between memory savings and usability by enforcing a 5 minute time limit on both the lifetime and location optimizations. Figure~\ref{fig:total_memory_savings} shows the overall reduction in peak memory usage achieved within these time limits. \ours{} achieves an average improvement of 30.4\% at batch size 1, and 36.1\% at batch size 32.

\paragraph{}

\begin{figure}
    \centering
    \includegraphics[width=0.48\textwidth]{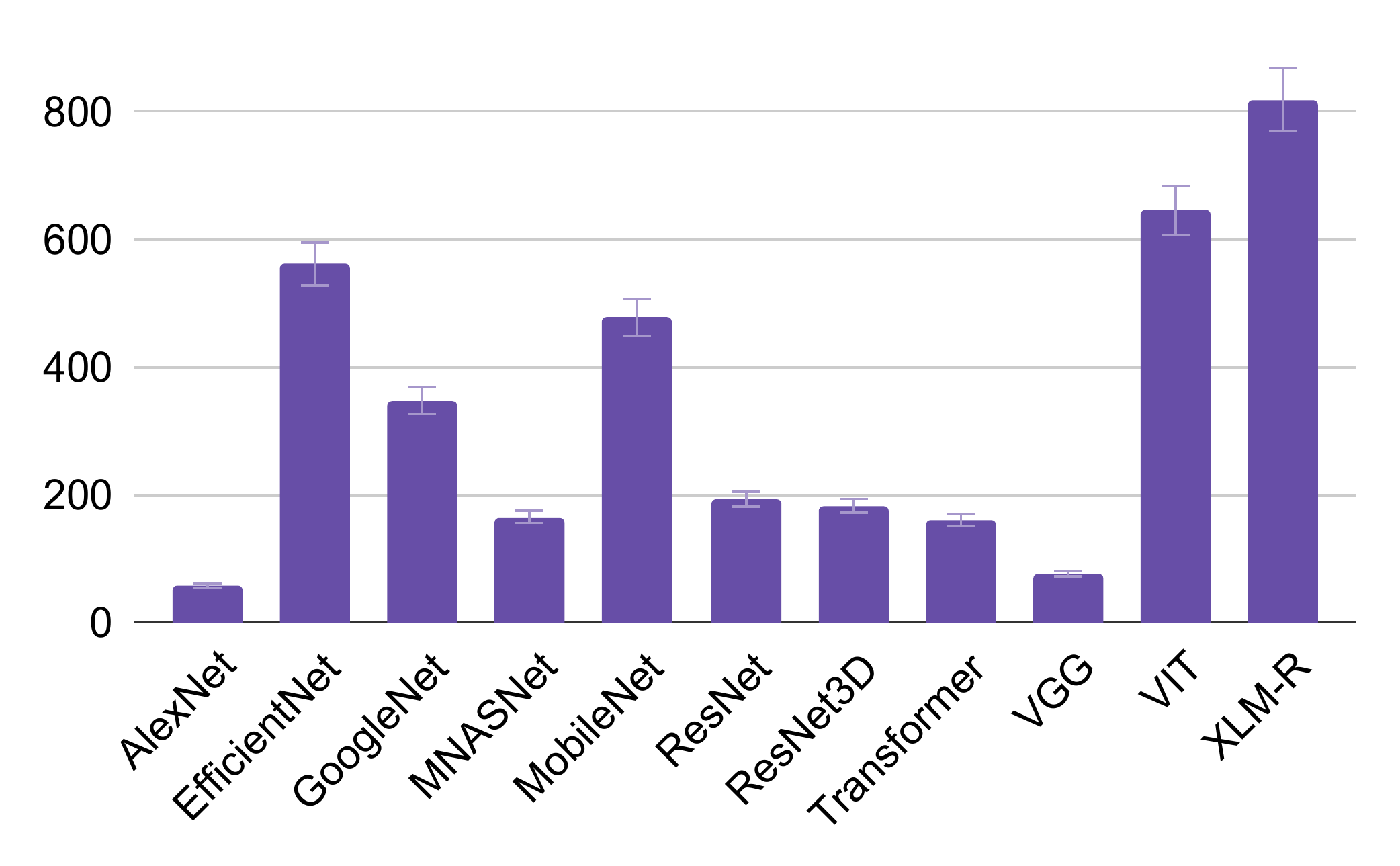}
    \caption{Runtime savings (in seconds) over PyTorch while training at batch size 32.}
    \label{fig:runtime_savings}
\end{figure}

PyTorch relies on dynamic memory allocation, which introduces some runtime overhead with each tensor allocation and deallocation. We compared the initial runtime penalty of our approach against the time the PyTorch allocator takes to manage memory while training neural networks at batch size 32. We assume that 1,000,000 iterations through the training loop are required to train a neural network, which is equivalent to processing the entire ImageNet~\cite{imagenet} dataset 25 times, and corresponds to 8 training epochs on the IWSLT2017 dataset~\cite{IWSLT2017}. Our approach proved to be slightly more runtime efficient overall, saving an average of 5 minutes (Figure ~\ref{fig:runtime_savings}).

\paragraph{}

Note that we didn't directly measure the total training time under both memory management schemes to reduce the noise in the measurements. Additionally, we didn't report our estimated runtime savings at batch size 1 since the premises of this use case, typically encountered on-device, are very different: while each user would only perform a few training iterations, our optimizations would be applied before shipping the application.

\section{Related Work}
\label{related_work}

Various approaches, complementary to ours, have been proposed to break the ``memory wall'' and train larger networks. The first technique distributes the computation for a single forward-backward iteration over several hardware devices, thus making more memory available overall. However, this approach, known as model parallelism~\cite{SageMakerModelParallelism}, significantly increases the financial cost of training deep neural networks since it requires access to additional expensive compute accelerators and fast networks. Furthermore, partitioning a deep neural network efficiently to balance communication and computation remains an open problem still actively researched~\cite{Gholami2018IntegratedMB, taso, Mirhoseini2018AHM}.

\paragraph{}

In parallel, the research community has developed numerous solutions to reduce the memory footprint of deep neural networks:
\begin{itemize}
    \item Novel neural network architectures reduce the number of parameters needed to achieve a given level of accuracy~\cite{iandola2017squeezenet, efficientnet}. Furthermore, automated search techniques known as neural architecture search~\cite{mnasnet} have been proposed to automatically design memory efficient models. The main drawbacks of these methods are that they are time consuming to deploy, and fail to match the result quality of state of the art DNNs.
    \item Model compression methods~\cite{MLSYS2020_d2ddea18} prune~\cite{weight_pruning, frankle_lottery,TaylorPruning,elkerdawy2022fire,He_2020_CVPR} or share~\cite{universal_transformer} weights to improve the efficiency of the model parameterization. However, the majority of these techniques require training the unpruned neural network first, and are therefore most useful for inference.
    \item Training using reduced precision arithmetic on 16-bit floating point or even quantized representations~\cite{fp8Training,INT8Training,bfloat16Training} significantly reduces memory~\cite{fan2020training,lin2016fixed}. However, these techniques can compromise the accuracy of the neural networks, make training unstable, and require careful implementation to be deployed successfully~\cite{MixedPrecisionTraining,Lin2016OvercomingCI}.
\end{itemize}

\paragraph{}

Several efforts have looked at the problem from a systems perspective, and presented solutions to reduce pressure on the memory subsystem. These techniques encompass:
\begin{itemize}
    \item In-memory tensor compression, which can result in minimal accuracy loss in many DNN applications~\cite{actnn, gist}. However, this comes with a runtime penalty, since the data must be compressed and uncompressed on the fly.
    \item Rematerialization, also known as checkpointing, discards activations in the forward pass to save memory, and recomputes those values as needed when computing the gradients. Numerous strategies to identify which activations to discard have been proposed~\cite{checkmate, echo, Chen2016TrainingDN, Griewank2000Algorithm7R, shah2021memory}. While effective at reducing memory usage, these techniques add extra computations, which increases the training time.
    \item Paging, aka spilling, consists of moving data between a small but high bandwidth and low latency memory pool, and a large but slow external memory. This has been demonstrated to effectively offload the data stored on a GPU device onto the host memory~\cite{Capuchin, autotm, SwapOutSwapIn}, but again increases training time due to extra memory transfers.
    \item More recently, combining several of these techniques has been proposed to increase their effectiveness and mitigate their drawbacks~\cite{beaumont:hal-03359793, poet} without fully eliminating them.
\end{itemize}

\paragraph{}

Additionally, some techniques developed primarily to increase execution speed are also beneficial for memory:
\begin{itemize}    
    \item Operator fusion can reduce memory footprint by avoiding the need to materialize large intermediate buffers and keep them around for backpropagation~\cite{DNNFusion}.
    \item Machine learning frameworks such as PyTorch~\cite{PyTorch} and TensorFlow~\cite{tensorflow} allow some of their operators to store the data they generate in one of their input tensors, thus avoiding the need to allocate an output tensor. This is known as in-place-update, and saves memory. However, users must manually modify their neural networks to leverage this capability, and it can lead to correctness issues if used indiscriminately~\cite{autodiff-pytorch}.
\end{itemize}

\paragraph{}

Optimizing the location of tensors in memory to reduce fragmentation, also known as the dynamic storage allocation problem, is NP-hard~\cite{dsa}. This problem has been studied in the context of deep learning by other researchers~\cite{profile_guided_mem_opt} who proposed an exact formulation to minimize the memory fragmentation of deep neural networks. However, their approach scaled poorly and only succeeded in optimizing two small neural networks in inference mode. As a result, they ultimately advocated for a heuristics based approach.

\paragraph{}

Improving the lifetime of tensors has also been studied before. Liberis \textit{et al.}~\cite{liberis2020neural} and Serenity~\cite{serenity} looked for a memory-optimal execution schedule by enumerating the topological orders of the DNN graph and calculating their peak memory usage. To speed things up, they both proposed dynamic programming based optimizations to prune the number of orderings they needed to consider. However, the complexity of their algorithms remains prohibitive at $O(|V|*2^{|V|})$ in both cases, and they only managed to make them work for inference on tiny graphs. Lin \textit{et al.}~\cite{on_device_training} also mentioned reordering computations as a way to enable operator fusion and reduce the peak memory footprint while training. Unfortunately, they didn't describe the algorithm they used to find a suitable node ordering.
\section{Conclusion}
The limited memory capacity of the hardware used by deep learning practitioners is one of the main challenges to train state-of-the art neural networks. This "memory wall" limits the size of the neural networks that can be trained, and ultimately impacts the quality of their predictions. Furthermore, as memory needs increase much faster than memory capacity, we expect this memory bottleneck to worsen over time.

\paragraph{}

To alleviate memory scarcity, we proposed to optimize both the lifetime and location of tensors in memory. We presented an ILP formulation of the problem, and extended it with ad-hoc simplifications to ensure that our approach scales to large neural networks.

\paragraph{}

We tested our solution, \ours{}, on a wide variety of neural networks. We demonstrated experimentally that it can locate tensors optimally in memory, thus eliminating the problem of memory fragmentation. Furthermore, we showed that it further decreases the peak memory usage of deep neural networks by optimizing the lifetime of the tensors, and, by combining these 2 techniques, \ours{} reduced peak memory usage by more than 30\% on average.

\paragraph{}

We also emphasized the practicality of \ours{}. We established empirically that it scales well and can handle large DNNs. We showed that it finds memory plans that are within 1\% of optimal in less than 10 minutes (and often in mere seconds), and measured that \ours{}'s optimization time is more than compensated for by eliminating the runtime overhead of dynamic allocation during training. 

\paragraph{}
\section{Acknowledgements}
We would like to thank Ana Klimovic and Foteini Strati, whose insightful comments and feedback helped improve the paper.

\bibliographystyle{plain}
\bibliography{references}

\begin{thebibliography}{10}

\bibitem{tensorflow}
Mart{\'\i}n Abadi, Paul Barham, Jianmin Chen, Zhifeng Chen, Andy Davis, Jeffrey
  Dean, Matthieu Devin, Sanjay Ghemawat, Geoffrey Irving, Michael Isard, et~al.
\newblock Tensorflow: A system for large-scale machine learning.
\newblock In {\em 12th $\{$USENIX$\}$ Symposium on Operating Systems Design and
  Implementation ($\{$OSDI$\}$ 16)}, pages 265--283, 2016.

\bibitem{serenity}
Byung~Hoon Ahn, Jinwon Lee, Jamie~Menjay Lin, Hsin-Pai Cheng, Jilei Hou, and
  Hadi Esmaeilzadeh.
\newblock Ordering chaos: Memory-aware scheduling of irregularly wired neural
  networks for edge devices, 2020.

\bibitem{asap_alap}
Zoltan Baruch.
\newblock Scheduling algorithms for high-level synthesis.
\newblock {\em ACAM Scientific Journal}, 5(1-2):48--57, 1996.

\bibitem{beaumont:hal-03359793}
Olivier Beaumont, Lionel Eyraud-Dubois, and Alena Shilova.
\newblock {Efficient Combination of Rematerialization and Offloading for
  Training DNNs}.
\newblock In {\em {NeurIPS 2021 - Thirty-fifth Conference on Neural Information
  Processing Systems}}, Virtual-only Conference, France, December 2021.

\bibitem{Bernstein1989OnTC}
David Bernstein, Michael Rodeh, and Izidor Gertner.
\newblock On the complexity of scheduling problems for parallel/pipelined
  machines.
\newblock {\em IEEE Trans. Computers}, 38:1308--1313, 1989.

\bibitem{MLSYS2020_d2ddea18}
Davis Blalock, Jose~Javier Gonzalez~Ortiz, Jonathan Frankle, and John Guttag.
\newblock What is the state of neural network pruning?
\newblock In I.~Dhillon, D.~Papailiopoulos, and V.~Sze, editors, {\em
  Proceedings of Machine Learning and Systems}, volume~2, pages 129--146, 2020.

\bibitem{ChallengesTransformerQuantization}
Yelysei Bondarenko, Markus Nagel, and Tijmen Blankevoort.
\newblock Understanding and overcoming the challenges of efficient transformer
  quantization.
\newblock In {\em Proceedings of the 2021 Conference on Empirical Methods in
  Natural Language Processing}, pages 7947--7969, 2021.

\bibitem{dag_schedule}
John Bruno and Ravi Sethi.
\newblock Code generation for a one-register machine.
\newblock {\em J. ACM}, 23(3):502–510, jul 1976.

\bibitem{RobustnessOverParameterization}
Sébastien Bubeck and Mark Sellke.
\newblock A universal law of robustness via isoperimetry.
\newblock In {\em NeurIPS 2021}, December 2021.

\bibitem{IWSLT2017}
Mauro Cettolo, Marcello Federico, Luisa Bentivogli, Jan Niehues, Sebastian
  St{\"u}ker, Katsuhito Sudoh, Koichiro Yoshino, and Christian Federmann.
\newblock Overview of the {IWSLT} 2017 evaluation campaign.
\newblock In {\em Proceedings of the 14th International Conference on Spoken
  Language Translation}, pages 2--14, Tokyo, Japan, December 14-15 2017.
  International Workshop on Spoken Language Translation.

\bibitem{actnn}
Jianfei Chen, Lianmin Zheng, Zhewei Yao, Dequan Wang, Ion Stoica, Michael
  Mahoney, and Joseph Gonzalez.
\newblock Actnn: Reducing training memory footprint via 2-bit activation
  compressed training.
\newblock In Marina Meila and Tong Zhang, editors, {\em Proceedings of the 38th
  International Conference on Machine Learning}, volume 139 of {\em Proceedings
  of Machine Learning Research}, pages 1803--1813. PMLR, 18--24 Jul 2021.

\bibitem{Semantic_Segmentation}
Liang-Chieh Chen, George Papandreou, Iasonas Kokkinos, Kevin~P. Murphy, and
  Alan~Loddon Yuille.
\newblock Semantic image segmentation with deep convolutional nets and fully
  connected crfs.
\newblock {\em CoRR}, abs/1412.7062, 2015.

\bibitem{Chen2016TrainingDN}
Tianqi Chen, Bing Xu, Chiyuan Zhang, and Carlos Guestrin.
\newblock Training deep nets with sublinear memory cost.
\newblock {\em ArXiv}, abs/1604.06174, 2016.

\bibitem{Chen2017Multiview3O}
Xiaozhi Chen, Huimin Ma, Ji~Wan, Bo~Li, and Tian Xia.
\newblock Multi-view 3d object detection network for autonomous driving.
\newblock {\em 2017 IEEE Conference on Computer Vision and Pattern Recognition
  (CVPR)}, pages 6526--6534, 2017.

\bibitem{sparse_transformers}
Rewon Child, Scott Gray, Alec Radford, and Ilya Sutskever.
\newblock Generating long sequences with sparse transformers.
\newblock {\em ArXiv}, abs/1904.10509, 2019.

\bibitem{xlm-r}
Alexis Conneau, Kartikay Khandelwal, Naman Goyal, Vishrav Chaudhary, Guillaume
  Wenzek, Francisco Guzm{\'{a}}n, Edouard Grave, Myle Ott, Luke Zettlemoyer,
  and Veselin Stoyanov.
\newblock Unsupervised cross-lingual representation learning at scale.
\newblock {\em CoRR}, abs/1911.02116, 2019.

\bibitem{levelization}
Wikipedia contributors.
\newblock Level structure.
\newblock \url{https://en.wikipedia.org/wiki/Level_structure}, 2022.
\newblock [Online; accessed 10-Oct-2022].

\bibitem{transformerXL}
Zihang Dai, Zhilin Yang, Yiming Yang, Jaime Carbonell, Quoc~V. Le, and Ruslan
  Salakhutdinov.
\newblock Transformer-xl: Attentive language models beyond a fixed-length
  context, 2019.

\bibitem{universal_transformer}
Mostafa Dehghani, Stephan Gouws, Oriol Vinyals, Jakob Uszkoreit, and Łukasz
  Kaiser.
\newblock Universal transformers, 2019.

\bibitem{imagenet}
Jia Deng, Wei Dong, Richard Socher, Li-Jia Li, Kai Li, and Li~Fei-Fei.
\newblock Imagenet: A large-scale hierarchical image database.
\newblock In {\em 2009 IEEE conference on computer vision and pattern
  recognition}, pages 248--255. Ieee, 2009.

\bibitem{bert}
Jacob Devlin, Ming-Wei Chang, Kenton Lee, and Kristina Toutanova.
\newblock Bert: Pre-training of deep bidirectional transformers for language
  understanding, 2018.

\bibitem{superresolution}
Chao Dong, Chen~Change Loy, Kaiming He, and Xiaoou Tang.
\newblock Image super-resolution using deep convolutional networks.
\newblock {\em IEEE Transactions on Pattern Analysis and Machine Intelligence},
  38:295--307, 2016.

\bibitem{vit}
Alexey Dosovitskiy, Lucas Beyer, Alexander Kolesnikov, Dirk Weissenborn,
  Xiaohua Zhai, Thomas Unterthiner, Mostafa Dehghani, Matthias Minderer, Georg
  Heigold, Sylvain Gelly, Jakob Uszkoreit, and Neil Houlsby.
\newblock An image is worth 16x16 words: Transformers for image recognition at
  scale, 2020.

\bibitem{elkerdawy2022fire}
Sara Elkerdawy, Mostafa Elhoushi, Hong Zhang, and Nilanjan Ray.
\newblock Fire together wire together: A dynamic pruning approach with
  self-supervised mask prediction.
\newblock In {\em Proceedings of the IEEE/CVF Conference on Computer Vision and
  Pattern Recognition (CVPR)}, June 2022.

\bibitem{nas}
Thomas Elsken, Jan~Hendrik Metzen, and Frank Hutter.
\newblock Neural architecture search: A survey.
\newblock {\em Journal of Machine Learning Research}, 20(55):1--21, 2019.

\bibitem{jemalloc}
Jason Evans.
\newblock jemalloc.
\newblock \url{https://github.com/jemalloc/jemalloc}.

\bibitem{fan2020training}
Angela Fan, Pierre Stock, Benjamin Graham, Edouard Grave, R{\'e}mi Gribonval,
  Herve Jegou, and Armand Joulin.
\newblock Training with quantization noise for extreme model compression.
\newblock {\em arXiv preprint arXiv:2004.07320}, 2020.

\bibitem{SlowFast}
Christoph Feichtenhofer, Haoqi Fan, Jitendra Malik, and Kaiming He.
\newblock Slowfast networks for video recognition.
\newblock In {\em Proceedings of the IEEE/CVF International Conference on
  Computer Vision (ICCV)}, October 2019.

\bibitem{frankle_lottery}
Jonathan Frankle and Michael Carbin.
\newblock The lottery ticket hypothesis: Finding sparse, trainable neural
  networks, 2018.
\newblock cite arxiv:1803.03635Comment: ICLR camera ready.

\bibitem{dsa}
Michael~R. Garey and David~S. Johnson.
\newblock Computers and intractability: A guide to the theory of
  np-completeness.
\newblock {\em Journal of Symbolic Logic}, 1979.

\bibitem{Gholami2018IntegratedMB}
Amir Gholami, Ariful Azad, Peter~H. Jin, Kurt Keutzer, and Aydın Buluç.
\newblock Integrated model, batch, and domain parallelism in training neural
  networks.
\newblock {\em Proceedings of the 30th on Symposium on Parallelism in
  Algorithms and Architectures}, 2018.

\bibitem{TCMalloc}
Google.
\newblock Tcmalloc.
\newblock \url{https://github.com/google/tcmalloc}.

\bibitem{Griewank2000Algorithm7R}
Andreas Griewank and Andrea Walther.
\newblock Algorithm 799: revolve: an implementation of checkpointing for the
  reverse or adjoint mode of computational differentiation.
\newblock {\em ACM Trans. Math. Softw.}, 26:19--45, 2000.

\bibitem{Gurobi}
{Gurobi Optimization, LLC}.
\newblock {Gurobi Optimizer Reference Manual}, 2022.

\bibitem{Hard2018FederatedLF}
Andrew Hard, Kanishka Rao, Rajiv Mathews, Françoise Beaufays, Sean Augenstein,
  Hubert Eichner, Chlo{\'e} Kiddon, and Daniel Ramage.
\newblock Federated learning for mobile keyboard prediction.
\newblock {\em ArXiv}, abs/1811.03604, 2018.

\bibitem{resnet}
Kaiming He, Xiangyu Zhang, Shaoqing Ren, and Jian Sun.
\newblock Deep residual learning for image recognition.
\newblock In {\em 2016 IEEE Conference on Computer Vision and Pattern
  Recognition (CVPR)}, pages 770--778, 2016.

\bibitem{He_2020_CVPR}
Yang He, Yuhang Ding, Ping Liu, Linchao Zhu, Hanwang Zhang, and Yi~Yang.
\newblock Learning filter pruning criteria for deep convolutional neural
  networks acceleration.
\newblock In {\em Proceedings of the IEEE/CVF Conference on Computer Vision and
  Pattern Recognition (CVPR)}, June 2020.

\bibitem{autotm}
Mark Hildebrand, Jawad Khan, Sanjeev Trika, Jason Lowe-Power, and Venkatesh
  Akella.
\newblock Autotm: Automatic tensor movement in heterogeneous memory systems
  using integer linear programming.
\newblock In {\em Proceedings of the Twenty-Fifth International Conference on
  Architectural Support for Programming Languages and Operating Systems},
  ASPLOS '20, page 875–890, New York, NY, USA, 2020. Association for
  Computing Machinery.

\bibitem{hochreiter1997lstm}
Sepp Hochreiter and J{\"u}rgen Schmidhuber.
\newblock Long short-term memory.
\newblock {\em Neural computation}, 9(8):1735--1780, 1997.

\bibitem{mobilenet}
Andrew~G. Howard, Menglong Zhu, Bo~Chen, Dmitry Kalenichenko, Weijun Wang,
  Tobias Weyand, Marco Andreetto, and Hartwig Adam.
\newblock Mobilenets: Efficient convolutional neural networks for mobile vision
  applications.
\newblock {\em ArXiv}, abs/1704.04861, 2017.

\bibitem{iandola2017squeezenet}
Forrest~N. Iandola, Song Han, Matthew~W. Moskewicz, Khalid Ashraf, William~J.
  Dally, and Kurt Keutzer.
\newblock Squeezenet: Alexnet-level accuracy with 50x fewer parameters and
  \ensuremath{<}0.5{MB} model size, 2017.

\bibitem{gist}
Animesh Jain, Amar Phanishayee, Jason Mars, Lingjia Tang, and Gennady
  Pekhimenko.
\newblock Gist: Efficient data encoding for deep neural network training.
\newblock In {\em Proceedings of the 45th Annual International Symposium on
  Computer Architecture}, ISCA '18, page 776–789. IEEE Press, 2018.

\bibitem{checkmate}
Paras Jain, Ajay Jain, Aniruddha Nrusimha, Amir Gholami, Pieter Abbeel, Joseph
  Gonzalez, Kurt Keutzer, and Ion Stoica.
\newblock Checkmate: Breaking the memory wall with optimal tensor
  rematerialization.
\newblock In I.~Dhillon, D.~Papailiopoulos, and V.~Sze, editors, {\em
  Proceedings of Machine Learning and Systems}, volume~2, pages 497--511, 2020.

\bibitem{taso}
Zhihao Jia, Sina Lin, Charles~R. Qi, and Alex Aiken.
\newblock Exploring hidden dimensions in accelerating convolutional neural
  networks.
\newblock In Jennifer Dy and Andreas Krause, editors, {\em Proceedings of the
  35th International Conference on Machine Learning}, volume~80 of {\em
  Proceedings of Machine Learning Research}, pages 2274--2283. PMLR, 10--15 Jul
  2018.

\bibitem{bfloat16Training}
Dhiraj~D. Kalamkar, Dheevatsa Mudigere, Naveen Mellempudi, Dipankar Das, Kunal
  Banerjee, Sasikanth Avancha, Dharma~Teja Vooturi, Nataraj Jammalamadaka,
  Jianyu Huang, Hector Yuen, Jiyan Yang, Jongsoo Park, Alexander Heinecke,
  Evangelos Georganas, Sudarshan Srinivasan, Abhisek Kundu, Misha Smelyanskiy,
  Bharat Kaul, and Pradeep Dubey.
\newblock A study of {BFLOAT16} for deep learning training.
\newblock {\em CoRR}, abs/1905.12322, 2019.

\bibitem{SageMakerModelParallelism}
Can Karakus, Rahul Huilgol, Fei Wu, Anirudh Subramanian, Cade Daniel, Derya
  {\c{C}}avdar, Teng Xu, Haohan Chen, Arash Rahnama, and Luis Quintela.
\newblock Amazon sagemaker model parallelism: {A} general and flexible
  framework for large model training.
\newblock {\em CoRR}, abs/2111.05972, 2021.

\bibitem{alexnet}
Alex Krizhevsky, Ilya Sutskever, and Geoffrey~E. Hinton.
\newblock Imagenet classification with deep convolutional neural networks.
\newblock {\em Communications of the ACM}, 60:84 -- 90, 2012.

\bibitem{liberis2020neural}
Edgar Liberis and Nicholas~D. Lane.
\newblock Neural networks on microcontrollers: saving memory at inference via
  operator reordering, 2020.

\bibitem{LostInPruning}
Lucas Liebenwein, Cenk Baykal, Brandon Carter, David Gifford, and Daniela Rus.
\newblock Lost in pruning: The effects of pruning neural networks beyond test
  accuracy.
\newblock In A.~Smola, A.~Dimakis, and I.~Stoica, editors, {\em Proceedings of
  Machine Learning and Systems}, volume~3, pages 93--138, 2021.

\bibitem{lin2016fixed}
Darryl Lin, Sachin Talathi, and Sreekanth Annapureddy.
\newblock Fixed point quantization of deep convolutional networks.
\newblock In {\em International conference on machine learning}, pages
  2849--2858. PMLR, 2016.

\bibitem{Lin2016OvercomingCI}
Darryl~Dexu Lin and Sachin~S. Talathi.
\newblock Overcoming challenges in fixed point training of deep convolutional
  networks.
\newblock {\em ArXiv}, abs/1607.02241, 2016.

\bibitem{on_device_training}
Ji~Lin, Ligeng Zhu, Wei-Ming Chen, Wei-Chen Wang, Chuang Gan, and Song Han.
\newblock On-device training under 256kb memory, 2022.

\bibitem{weight_pruning}
Christos Louizos, Max Welling, and Diederik~P. Kingma.
\newblock Learning sparse neural networks through $l_0$ regularization, 2018.

\bibitem{SwapOutSwapIn}
Chen Meng, Minmin Sun, Jun Yang, Minghui Qiu, and Yang Gu.
\newblock Training deeper models by gpu memory optimization on tensorflow.
\newblock In {\em NIPS 2017 Workshop on ML Systems}, 2017.

\bibitem{Mirhoseini2018AHM}
Azalia Mirhoseini, Anna Goldie, Hieu Pham, Benoit Steiner, Quoc~V. Le, and Jeff
  Dean.
\newblock A hierarchical model for device placement.
\newblock In {\em ICLR}, 2018.

\bibitem{TaylorPruning}
Pavlo Molchanov, Arun Mallya, Stephen Tyree, Iuri Frosio, and Jan Kautz.
\newblock Importance estimation for neural network pruning.
\newblock In {\em Proceedings of the IEEE Conference on Computer Vision and
  Pattern Recognition}, 2019.

\bibitem{OscilationsQAT}
Markus Nagel, Marios Fournarakis, Yelysei Bondarenko, and Tijmen Blankevoort.
\newblock Overcoming oscillations in quantization-aware training.
\newblock In Kamalika Chaudhuri, Stefanie Jegelka, Le~Song, Csaba Szepesvari,
  Gang Niu, and Sivan Sabato, editors, {\em Proceedings of the 39th
  International Conference on Machine Learning}, volume 162 of {\em Proceedings
  of Machine Learning Research}, pages 16318--16330. PMLR, 17--23 Jul 2022.

\bibitem{DNNFusion}
Wei Niu, Jiexiong Guan, Yanzhi Wang, Gagan Agrawal, and Bin Ren.
\newblock Dnnfusion: Accelerating deep neural networks execution with advanced
  operator fusion.
\newblock In {\em Proceedings of the 42nd ACM SIGPLAN International Conference
  on Programming Language Design and Implementation}, PLDI 2021, page
  883–898, New York, NY, USA, 2021. Association for Computing Machinery.

\bibitem{MixedPrecisionTraining}
NVidia.
\newblock Mixed precision training.
\newblock
  \url{https://docs.nvidia.com/deeplearning/performance/mixed-precision-training/index.html}.
\newblock [Online; accessed 13-Oct-2022].

\bibitem{autodiff-pytorch}
Adam Paszke, Sam Gross, Soumith Chintala, Gregory Chanan, Edward Yang, Zachary
  DeVito, Zeming Lin, Alban Desmaison, Luca Antiga, and Adam Lerer.
\newblock Automatic differentiation in pytorch.
\newblock In {\em NIPS 2017 Workshop on Autodiff}, 2017.

\bibitem{PyTorch}
Adam Paszke, Sam Gross, Francisco Massa, Adam Lerer, James Bradbury, Gregory
  Chanan, Trevor Killeen, Zeming Lin, Natalia Gimelshein, Luca Antiga, Alban
  Desmaison, Andreas Kopf, Edward Yang, Zachary DeVito, Martin Raison, Alykhan
  Tejani, Sasank Chilamkurthy, Benoit Steiner, Lu~Fang, Junjie Bai, and Soumith
  Chintala.
\newblock Pytorch: An imperative style, high-performance deep learning library.
\newblock In {\em Advances in Neural Information Processing Systems 32}, pages
  8024--8035. Curran Associates, Inc., 2019.

\bibitem{poet}
Shishir~G. Patil, Paras Jain, Prabal Dutta, Ion Stoica, and Joseph Gonzalez.
\newblock {POET}: Training neural networks on tiny devices with integrated
  rematerialization and paging.
\newblock In Kamalika Chaudhuri, Stefanie Jegelka, Le~Song, Csaba Szepesvari,
  Gang Niu, and Sivan Sabato, editors, {\em Proceedings of the 39th
  International Conference on Machine Learning}, volume 162 of {\em Proceedings
  of Machine Learning Research}, pages 17573--17583. PMLR, 17--23 Jul 2022.

\bibitem{paulik}
Matthias Paulik, Matt Seigel, Henry Mason, Dominic Telaar, Joris Kluivers,
  Rogier van Dalen, Chi~Wai Lau, Luke Carlson, Filip Granqvist, Chris
  Vandevelde, Sudeep Agarwal, Julien Freudiger, Andrew Byde, Abhishek Bhowmick,
  Gaurav Kapoor, Si~Beaumont, Áine Cahill, Dominic Hughes, Omid Javidbakht,
  Fei Dong, Rehan Rishi, and Stanley Hung.
\newblock Federated evaluation and tuning for on-device personalization: System
  design and applications, 2021.

\bibitem{Capuchin}
Xuan Peng, Xuanhua Shi, Hulin Dai, Hai Jin, Weiliang Ma, Qian Xiong, Fan Yang,
  and Xuehai Qian.
\newblock Capuchin: Tensor-based gpu memory management for deep learning.
\newblock In {\em Proceedings of the Twenty-Fifth International Conference on
  Architectural Support for Programming Languages and Operating Systems},
  ASPLOS '20, page 891–905, New York, NY, USA, 2020. Association for
  Computing Machinery.

\bibitem{profile_guided_mem_opt}
Taro Sekiyama, Takashi Imamichi, Haruki Imai, and Rudy Raymond.
\newblock Profile-guided memory optimization for deep neural networks, 2018.

\bibitem{dnn_trends}
Jaime Sevilla, Lennart Heim, Anson Ho, Tamay Besiroglu, Marius Hobbhahn, and
  Pablo Villalobos.
\newblock Compute trends across three eras of machine learning.
\newblock {\em 2022 International Joint Conference on Neural Networks (IJCNN)},
  Jul 2022.

\bibitem{shah2021memory}
Aashaka Shah, Chao-Yuan Wu, Jayashree Mohan, Vijay Chidambaram, and Philipp
  Kraehenbuehl.
\newblock Memory optimization for deep networks.
\newblock In {\em International Conference on Learning Representations}, 2021.

\bibitem{vgg}
Karen Simonyan and Andrew Zisserman.
\newblock Very deep convolutional networks for large-scale image recognition.
\newblock In Yoshua Bengio and Yann LeCun, editors, {\em 3rd International
  Conference on Learning Representations, {ICLR} 2015, San Diego, CA, USA, May
  7-9, 2015, Conference Track Proceedings}, 2015.

\bibitem{increaseTheBatchSize}
Samuel~L. Smith, Pieter-Jan Kindermans, and Quoc~V. Le.
\newblock Don't decay the learning rate, increase the batch size.
\newblock In {\em International Conference on Learning Representations}, 2018.

\bibitem{googlenet}
Christian Szegedy, Wei Liu, Yangqing Jia, Pierre Sermanet, Scott Reed, Dragomir
  Anguelov, Dumitru Erhan, Vincent Vanhoucke, and Andrew Rabinovich.
\newblock Going deeper with convolutions.
\newblock In {\em 2015 IEEE Conference on Computer Vision and Pattern
  Recognition (CVPR)}, pages 1--9, 2015.

\bibitem{Tai_2017_CVPR}
Ying Tai, Jian Yang, and Xiaoming Liu.
\newblock Image super-resolution via deep recursive residual network.
\newblock In {\em Proceedings of the IEEE Conference on Computer Vision and
  Pattern Recognition (CVPR)}, July 2017.

\bibitem{mnasnet}
Mingxing Tan, Bo~Chen, Ruoming Pang, Vijay Vasudevan, Mark Sandler, Andrew
  Howard, and Quoc~V. Le.
\newblock Mnasnet: Platform-aware neural architecture search for mobile, 2019.

\bibitem{efficientnet}
Mingxing Tan and Quoc~V. Le.
\newblock Efficientnet: Rethinking model scaling for convolutional neural
  networks.
\newblock {\em ArXiv}, abs/1905.11946, 2019.

\bibitem{resnet3d}
Du~Tran, Heng Wang, Lorenzo Torresani, Jamie Ray, Yann LeCun, and Manohar
  Paluri.
\newblock A closer look at spatiotemporal convolutions for action recognition.
\newblock In {\em 2018 IEEE/CVF Conference on Computer Vision and Pattern
  Recognition}, pages 6450--6459, 2018.

\bibitem{transformer}
Ashish Vaswani, Noam Shazeer, Niki Parmar, Jakob Uszkoreit, Llion Jones,
  Aidan~N Gomez, \L~ukasz Kaiser, and Illia Polosukhin.
\newblock Attention is all you need.
\newblock In I.~Guyon, U.~Von Luxburg, S.~Bengio, H.~Wallach, R.~Fergus,
  S.~Vishwanathan, and R.~Garnett, editors, {\em Advances in Neural Information
  Processing Systems}, volume~30. Curran Associates, Inc., 2017.

\bibitem{fp8Training}
Naigang Wang, Jungwook Choi, Daniel Brand, Chia-Yu Chen, and Kailash
  Gopalakrishnan.
\newblock Training deep neural networks with 8-bit floating point numbers.
\newblock In S.~Bengio, H.~Wallach, H.~Larochelle, K.~Grauman, N.~Cesa-Bianchi,
  and R.~Garnett, editors, {\em Advances in Neural Information Processing
  Systems}, volume~31. Curran Associates, Inc., 2018.

\bibitem{echo}
Bojian Zheng, Nandita Vijaykumar, and Gennady Pekhimenko.
\newblock Echo: Compiler-based gpu memory footprint reduction for lstm rnn
  training.
\newblock In {\em 2020 ACM/IEEE 47th Annual International Symposium on Computer
  Architecture (ISCA)}, pages 1089--1102, 2020.

\bibitem{INT8Training}
Feng Zhu, Ruihao Gong, Fengwei Yu, Xianglong Liu, Yanfei Wang, Zhelong Li,
  Xiuqi Yang, and Junjie Yan.
\newblock Towards unified int8 training for convolutional neural network.
\newblock In {\em IEEE/CVF Conference on Computer Vision and Pattern
  Recognition (CVPR)}, June 2020.

\end{thebibliography}

\end{document}